%% file: main_paper.tex
\documentclass[10pt,twocolumn,letterpaper]{article}

\usepackage[pagenumbers]{cvpr} 

\usepackage{graphicx}
\usepackage{amsmath}
\usepackage{amssymb}
\usepackage{booktabs}
\usepackage[titletoc]{appendix}

\usepackage[accsupp]{axessibility}

\DeclareMathOperator*{\argmin}{arg\,min}

\usepackage[pagebackref,breaklinks,colorlinks]{hyperref}

\usepackage[capitalize]{cleveref}
\crefname{section}{Sec.}{Secs.}
\Crefname{section}{Section}{Sections}
\Crefname{table}{Table}{Tables}
\crefname{table}{Tab.}{Tabs.}

\def\cvprPaperID{5212} 
\def\confName{CVPR}
\def\confYear{2023}

\begin{document}

\title{Revisiting Multimodal Representation in Contrastive Learning: From Patch and Token Embeddings to Finite Discrete Tokens}

\author{
    Yuxiao Chen\textsuperscript{\rm 1\thanks{This work was done during a research internship at ByteDance.}},
    Jianbo Yuan\textsuperscript{\rm 2},
    Yu Tian\textsuperscript{\rm 2},
    Shijie Geng\textsuperscript{\rm 1,2},
    Xinyu Li\textsuperscript{\rm 2}, \\
    Ding Zhou\textsuperscript{\rm 2},
    Dimitris N. Metaxas\textsuperscript{\rm 1 \thanks{Dimitris N. Metaxas has been supported by NSF IUCRC CARTA-1747778, 2235405, 2212301, 1951890, 2003874.}},
    Hongxia Yang\textsuperscript{\rm 3} \\
    \textsuperscript{\rm 1}Rutgers University \quad
    \textsuperscript{\rm 2}ByteDance Inc. \quad
    \textsuperscript{\rm 3}Zhejiang University \\
    \texttt{\{yc984, sg1309, dnm\}@rutgers.edu},\\ \texttt{\{jianbo.yuan, yutian.yt, lixinyu.arthur, ding.zhou\}@bytedance.com} \\
    \texttt{hongxia.yang1@gmail.com} 
}
\maketitle
\begin{abstract}
\vspace{-12pt}
Contrastive learning-based vision-language pre-training approaches, such as CLIP, have demonstrated great success in many vision-language tasks. These methods achieve cross-modal alignment by encoding a matched image-text pair with similar feature embeddings, which are generated by aggregating information from visual patches and language tokens. However, direct aligning cross-modal information using such representations is challenging, as visual patches and text tokens differ in semantic levels and granularities. To alleviate this issue, we propose a Finite Discrete Tokens (FDT) based multimodal representation. FDT is a set of learnable tokens representing certain visual-semantic concepts. Both images and texts are embedded using shared FDT by first grounding multimodal inputs to FDT space and then aggregating the activated FDT representations. The matched visual and semantic concepts are enforced to be represented by the same set of discrete tokens by a sparse activation constraint. As a result, the granularity gap between the two modalities is reduced. Through both quantitative and qualitative analyses, we demonstrate that using FDT representations in CLIP-style models improves cross-modal alignment and performance in visual recognition and vision-language downstream tasks.  Furthermore, we show that our method can learn more comprehensive representations, and the learned FDT capture meaningful cross-modal correspondence, ranging from objects to actions and attributes.\footnote{The source code can be found
at \textcolor{magenta}{\url{https://github.com/yuxiaochen1103/FDT}}.} 

\end{abstract}

\vspace{-12pt}
\section{Introduction}
\label{sec:intro}
\input{sections/introduction_v2.tex}

\begin{figure*}[t]
\begin{center}
   \includegraphics[width=0.9\linewidth]{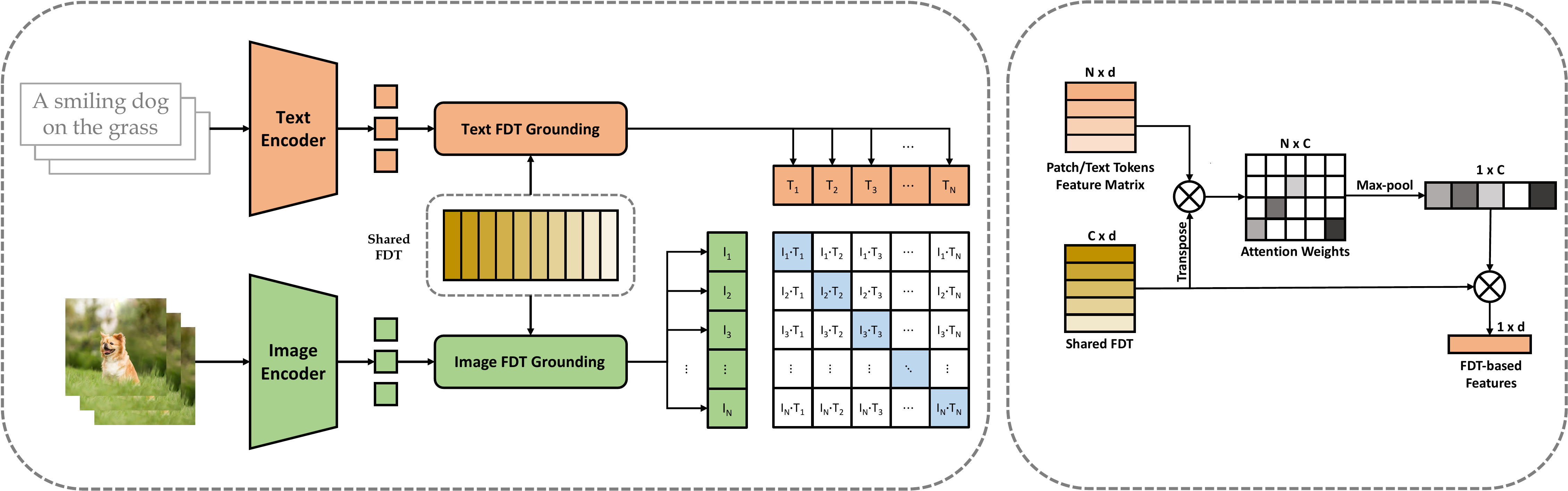}
\end{center}
\caption{\textbf{Left:} Overview of the proposed method. Both the image and text information is encoded with shared FDT during cross-modal contrastive pre-training. \textbf{Right:} The process of grounding image or text features to FDT. The attention weights between visual patch/language token and FDT are first calculated, and then max-pooled over all visual patches/language tokens. The attention-weighted sum of FDT is calculated as the FDT-based features. }
\vspace{-10pt}
\label{fig:method}
\end{figure*}

\section{Related Work}
\label{sec:rw}
\input{sections/related_work.tex}

\section{Method}
\label{sec:method}

\input{sections/method.tex}

\section{Experiments}
\label{sec:exp}
\input{sections/experiment_v2.tex}

{\small
\bibliographystyle{ieee_fullname}
\bibliography{egbib}
}

\clearpage
\appendix
\input{sections/supp_final.tex}

\end{document}

%% file: sections/introduction_v2.tex
\begin{figure}[t]
\begin{center}
    \includegraphics[width=\linewidth]{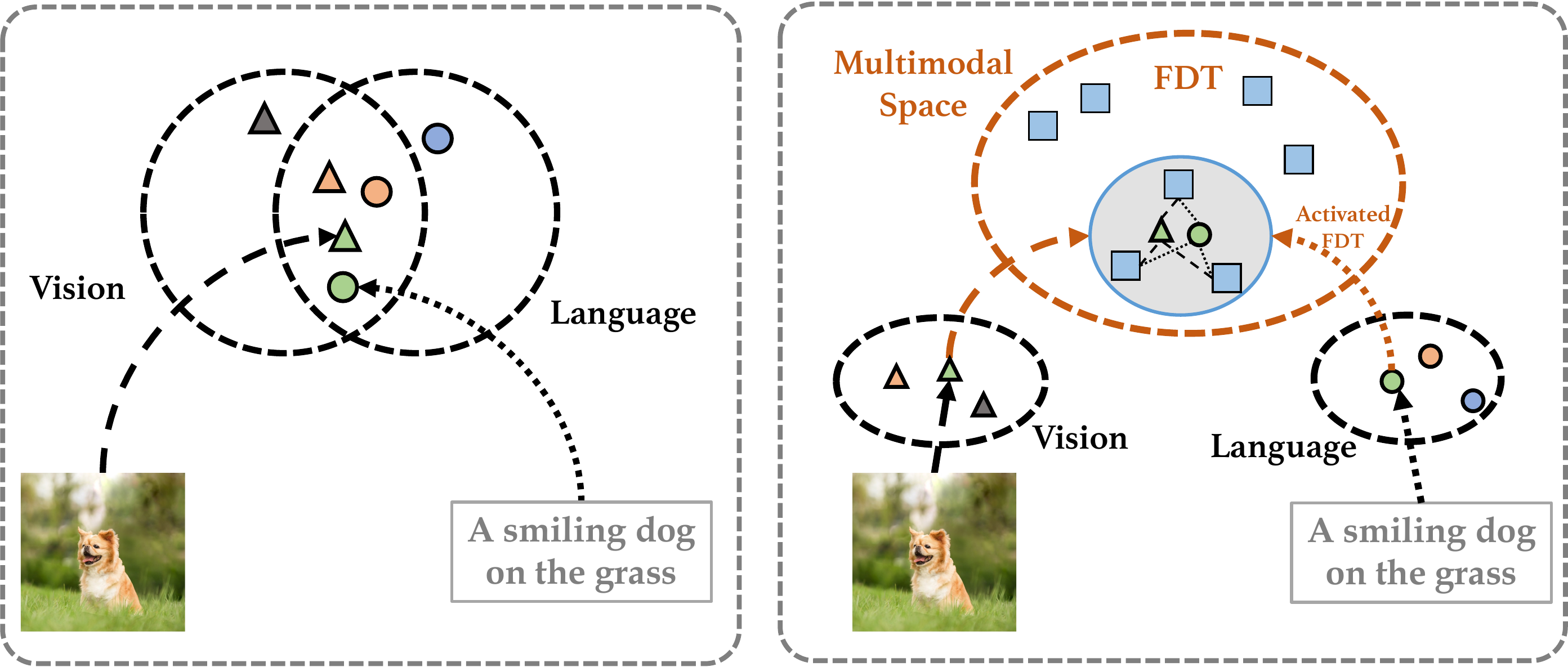}
\end{center}
\vspace{-10pt}
\caption{Comparison of different feature representation learning methods. \textbf{Left}: contrastive vision-language pre-training (CLIP). \textbf{Right}: CLIP with our proposed finite discrete tokens (FDT).}
\vspace{-10pt}
\label{fig:logic_flow}
\end{figure}

\vspace{-5pt}
Recently, the Contrastive Language-Image Pre-training (CLIP) framework \cite{clip, align} has demonstrated notable capabilities for learning powerful and transferable feature representations~\cite{zhang2022learning,zhang2023prompt,zhang2023parameter,zhang2022sine,gao2021clip,lin2022frozen}.~In this framework, models are trained to align text and image information in a two-stream approach where image and text representations are extracted through two separate encoders. The InfoNCE loss \cite{clip} is used to train the encoders which enforces the representations of matched image-text pairs to be closer, while those of unmatched pairs to be far apart (as shown in Figure \ref{fig:logic_flow} (Left)).

However, the fact that the information conveyed in images and text captions is naturally of different levels of granularities \cite{Gra_unif, image_infinity} is not considered by such models. For example, an image of a dog also portrays various lower-level attributes, such as its breed, fur color, body size, and shape, while the textual description, such as ``a smiling dog'', is generally more abstract and compact. In CLIP, images and text captions are represented through the aggregation of visual patches and text tokens without explicitly aligning the visual and semantic concepts at the same level of granularity. It can cause challenges in multimodal representation learning, or even potentially result in performance degradation \cite{tcl}. Additionally, the learned models may overlook certain semantic concepts\cite{learning_concepts}.
Therefore, we argue that unifying the information granularities of images and texts can help generate better multimodal representations.

In this paper, we propose a new \textbf{F}inite \textbf{D}iscrete \textbf{T}okens (FDT) based representations.~FDT is a set of \emph{learnable tokens} that encode cross-modal shared semantic concepts. Both image and text are represented as the combinations of FDT shared between modalities so that the information granularities are unified (see Figure \ref{fig:logic_flow} (Right)). Figure \ref{fig:method} gives an overview of our method. For an image, its patch embeddings are first extracted by an image encoder. The correspondence between the FDT and the image is then measured by max pooling over the attention weights of FDT among all patches. Finally, the FDT-based representation of the image is calculated as the attention-weighted sum of FDT. The FDT-based embeddings for input texts can be constructed in the same way. 
The encoders and FDT are trained to pull close the FDT-based representations of matched image-text pairs while pushing away those of unmatched pairs by using the InfoNCE loss. 
To the point of leveraging a shared FDT across modalities is to enforce the matched visual and semantic concepts to be represented by the same discrete tokens. For example, the visual patches of a dog and the word ``dog'' should activate the same subsets of FDT. We empirically demonstrate that this can be achieved by simply enforcing relatively sparse attention-weights between FDT and the inputs.

We conduct extensive experiments covering a wide range of pre-training settings and downstream tasks to evaluate the proposed method.~We conclude with the following key observations:~
(1) Our approach exhibits consistent performance enhancements across various pre-training dataset scales, CLIP-based pre-training frameworks \cite{declip}, and encoder architectures. 
Notably, our method outperforms CLIP by 5.0\% on zero-shot image classification when pre-training on 145M datasets, and by 33.4\% in image-text retrieval with 30M datasets; (2) Our method tends to alleviate the model degradation problem and learns more comprehensive feature representations than CLIP; (3) The learned FDT exhibit better: we visualize FDT's correspondent patches and language tokens, and the results show that FDT successfully capture and align visual-semantic concepts including objects, attributes, and actions.

%% file: sections/related_work.tex
\noindent\textbf{Vision and Language Pre-training.} Vision and language pre-training methods can be briefly classified into two-stream 
and single-stream 
models based on their architectures.~A typical two-stream model leverages individual encoders to extract continuous feature embeddings from the inputs, and enforces the embeddings of a matched image-text pair to be similar by using contrastive learning \cite{clip,align,geng2023hiclip} and additional self-supervised tasks \cite{declip,yang2022vision}. Inherited from the encoder design, these feature embeddings convey information aggregated from local vision patches and language tokens, which encompass different semantic levels and granularities and are constrained by how patches are generated. Therefore, we propose FDT-based representations to directly perform contrastive learning on FDT that denotes high-level vision-semantic concepts.~The single-stream approaches feed all inputs together into a unified encoder (mostly transformers) to enhance the cross-modal interactions for a better cross-modal alignment \cite{ALBEF,li2022unimo,singh2022flava,yang2022vision,chen2023more,chen2018you}. For simplicity, we also clarify models consisting of individual encoders followed by multimodal fusion operations (late-fusion) as one-stream, because it requires the inputs from all modalities for inference and hence does not support ANN, 
similar to a typical one-stream model (early-fusion). To combine the best of both worlds, FDT-based representations bridge the gap between different modalities with cross-modal interactions by vision-to-token and language-to-token information exchange, while maintaining a two-stream structure.

\noindent\textbf{Vector-Quantization and Codebook.} 
Vector-quantization is first proposed for image generation showing that image information can be encoded by discrete representations (namely \emph{codebook}) \cite{vqvae}.~Each image patch is represented by its nearest-neighbor code's embedding, and the decoder reconstructs the input image based on these code embeddings.~Because finding \emph{nearest-neighbor} is non-differentiable, the codebook is trained either by minimizing the distance between the code and image patch embeddings when the encoder is stop-gradient, or by exponential moving averages (EMA). Applying VQ to multimodal pre-training is more challenging, as the codebook now needs to accommodate multimodal contents and is often found to be sensitive to initialization (cold-start problem).~To address these challenges, previous studies leverage encoder or code warm-up \cite{crossdiscrete}, knowledge distilled vision tokenizers from pre-trained vision-language models \cite{peng2022beit2}, one-stream models to enforce multimodal code learning \cite{soho,li2022unimo}, and a combination of these techniques \cite{wang2022beit3}. As a comparison, our approach is designed to be more intuitive where only differentiable operations are used and it can be trained end2end from scratch while still maintaining a two-stream structure for ANN in large-scale retrieval tasks. More technical details will be discussed in Section \ref{sec:method:code}.

\noindent\textbf{Dictionary Learning.} Dictionary learning is another group of discrete representation learning in addition to VQ \cite{mm_dict_cls,mm_dict_conv,conv_dict}. Given a dictionary matrix \cite{conv_dict}, the representation of a signal is the weights that can linearly combine the dictionary matrix to reconstruct the signal with minimal error. When learning multi-modal representations \cite{mm_dict_cls,mm_dict_conv}, a shared dictionary matrix is used for facilitating cross-modal information alignment and fusion. The dictionary is served as the cross-modal information anchor, which shares the same idea as our method. However, the models are trained to solve a slow optimization problem, and the feature learned by solving the reconstruction or generative problem may have limited discriminative capability. By contrast, our model is trained end-to-end to learn discriminative information.

%% file: sections/method.tex

\subsection{Revisiting Feature Representations in CLIP}
\label{sec:clip}

In CLIP, the image and text features are the aggregation of the embeddings of image patches or language tokens, respectively. Specifically, the image encoder takes an image as input and extracts the patch or local region embeddings based on the self-attention \cite{vit}, or convolution operations \cite{resnet}. The obtained patch features are then aggregated as the final representation of the image $f_{v}$ by using the attention pooling or the [CLS] token \cite{clip,vit}, which can be formulated as: 
\vspace{-5pt}
\begin{align}
    \label{patch_att}
    w_{p_i} = \frac{e^{<f_{g},f_{p_i}>}}
    {\sum_{j}^{N_v}e^{<f_{g},f_{p_j}>}}, \\
    f_{v} = \sum_{i} ^ {N_v} (w_{p_i} \cdot f_{p_i}). 
\end{align}
Here, $w_{p_i}$ is the weight of $i$-th patch, which measures the importance of the patch to the final representation. $<,>$ is the inner-product function. $N_v$ is the number of patches, and $f_{p_i}$ denotes the embedding of $i$-th patch. $f_{g}$ is the [CLS] token embedding or the average-pooled patch embedding, which embeds the global image information.






Similarly, for the text encoder, the extracted text representation of an input sentence can also be regarded as the weighted sum of language token embeddings:

\begin{equation}
\label{clip_t}
f_{t} = \sum_{i} ^ {N_t} (w_{t_i} \cdot f_{t_i}),
\end{equation}
where $N_t$ is the number of language tokens. $f_{t_i}$ is the embedding of the $i$-th language token. It is extracted with the self-attention operations \cite{bert,gpt}, which model the relationship among the language tokens. $w_{t_i}$ is the weight of the  $i$-th language token, which is calculated by the following Equation \ref{patch_att} using the text [CLS] token.




Equations~\ref{patch_att} and \ref{clip_t} suggest that images or texts are represented by two different bases: visual patches and language tokens. However, the information conveyed by image patches and language tokens may have different semantic meanings and granularities. Additionally, the bases are dynamic, since the visual patches or language tokens of different images or texts are different. It may increase the difficulty of learning an optimal alignment between image and text features \cite{tcl,learning_concepts}. Thus, the encoders may fail to capture important semantic concepts shared in both modalities and may encode irrelevant information.




\subsection{FDT-based Representation}
\label{sec:method:code}
To address the aforementioned limitations of feature representation in CLIP, we propose the FDT-based representation. Figure \ref{fig:method} gives an overview of our proposed method. Instead of representing the image and text with different bases, FDT serve as the common bases for both the image and text representations. As a result, the granularities of cross-modal information are explicitly unified. Moreover, the FDT encode the semantic information shared by both modalities. It can be regarded as prior knowledge that guides image and text encoders to extract feature embeddings.
In the following, we elaborate on the steps necessary to achieve FDT-based representations:

\vspace{0.1cm}
\noindent{\textbf{Grounding to FDT.}} Let $\{c_i|i=1,..., C\}$ be FDT, where $C$ is the number of shared tokens, and $c_i$ is the $i$-th discrete token. Given an input image, its patch embeddings are first extracted using the image encoder. The extracted patch embeddings are then projected to the FDT space by using a projecting function. The relevance between the image and a token is obtained by calculating the inner product between the projected patch embeddings and the token, and selecting the maximal value, which can be formulated as
\begin{equation}
    \label{eq:img_relevance}
    r_i^v = \underset{j}{\mathrm{max}}<f_{p_j}, c_i>,
\end{equation}
where $r_i^v$ is the relevance between the image and the $i$-th tokens. Intuitively, the proposed patch-level relevance calculation mechanism may enjoy two advantages: (1) it can capture small objects that exist in a single patch; (2) it helps remove the influence of irrelevant noisy patches that have low relevance to all FDT.

The relevance between the image and FDT is normalized by a Softmax function, which generates the final weights of each token  as follows:
\vspace{-5pt}
\begin{equation}
    \label{eq:img_softmax}
    w^v_i = \frac{e^{r_i^v}}
    {\sum_{j}^C{e^{r_j^v}}},
\end{equation} where $w^v_i$ is the weight of the $i$-th token with respect to the image.  Similarly, the weight $w^t_i$ of the $i$-th token assigned by an input text can be calculated using
\vspace{-5pt}
\begin{equation}
    \label{eq:text_relevance}
    r_i^t = \underset{j}{\mathrm{max}}<f_{t_j}, c_i>,
\end{equation}
\begin{equation}
    \label{eq:text_softmax}
    w^t_i = \frac{e^{r_i^t}}
    {\sum_{j}^C{e^{r_j^t}}}.
\end{equation}

Intuitively, FDT can be treated as prior knowledge for the image or text information. With the help of FDT, the extracted features of both modalities are grounded to a shared manifold space, thus enabling the cross-modal interaction.



\vspace{0.1cm}
\noindent{\bf Normalizing Concept Weights with Sparse Constraints.} We expect the normalized weights of FDT to be sparse, since it can largely reduce noise and make the results more interpretable \cite{conv_dict, mm_dict_cls}. Additionally, we empirically show that sparsity is crucial for FDT to learn cross-modal correspondence, where a token corresponds to the same image and text semantic meaning. We use the Sparsemax function \cite{sparsemax} for sparser weights, which is defined as:
\begin{equation}
    \label{eq:sparsemax}
    \argmin_{\pmb{p}\in\Delta^{K-1}}{\| \pmb{p}- \pmb{r} \|^2},
\end{equation}
where $\pmb{r}$ is the vector consisting of the relevance score between the image or text and FDT (Equation \ref{eq:img_relevance} and \ref{eq:text_relevance}). This function first calculates a threshold, and then sets the weights below the threshold to zero for sparsity. In contrast, the commonly used Softmax function cannot explicitly assign FDT with exactly zero probabilities.


\vspace{0.1cm}
\noindent{\textbf{Generating FDT-based Embeddings.~}} Given the normalized weights, the FDT-based features of the image $f^{\mathrm{FDT}}_v$ and text $f^{\mathrm{FDT}}_t$ are the weighted sum of FDT:
\vspace{-5pt}
\begin{equation}
\label{fdt_v}
    f^{\mathrm{FDT}}_v = \sum_{i}^{C} w_i^v \cdot c_i
\end{equation}
\begin{equation}
\label{fdt_t}
     f^{\mathrm{FDT}}_t = \sum_{i}^{C} w_i^t \cdot c_i 
\end{equation}
Equations \ref{fdt_v} and \ref{fdt_t} show that image and text features are represented by the same base FDT, which explicitly unifies the granularities of image and text information.


Given the FTD-based features, the encoders and FDT are trained to make the similarity between FDT-based features of matched image-text pairs larger than those of unmatched pairs:
\begin{align}
    \label{eq:infonce}
    \mathcal{L} = -\frac{1}{N} \sum_{i}^{N} \log\frac{\mathrm{exp}\left(\mathrm{sim}\left(f^{\mathrm{FDT}}_{v_i}, f^{\mathrm{FDT}}_{t_i}\right) / \tau\right)}{\sum_{j=1}^{N} \mathrm{exp}\left(\mathrm{sim}\left(f^{\mathrm{FDT}}_{v_i}, f^{\mathrm{FDT}}_{t_j}\right) / \tau\right)} \nonumber \\ -\frac{1}{N} \sum_{i}^{N}\log\frac{\mathrm{exp}\left(\mathrm{sim}\left(f^{\mathrm{FDT}}_{t_i}, f^{\mathrm{FDT}}_{v_i}\right) / \tau\right)}{\sum_{j=1}^{N} \mathrm{exp}\left(\mathrm{sim}\left(f^{\mathrm{FDT}}_{t_i}, f^{\mathrm{FDT}}_{v_j}\right) / \tau\right)},
\end{align} 
where $N$ is the number of matched image-text pairs, $sim$ is the cosine similarity function, and $\tau$ is the temperature hyper-parameter.

Intuitively, the equation shows that FDT are updated based on both the image and text modalities, and thus FDT is trained to learn the information shared by both modalities.

%% file: sections/experiment_v2.tex
\subsection{Experimental Settings}
\noindent{\textbf{Pre-training Datasets.~}} We use four publicly available datasets, including YFCC-15M V2 \cite{declip_benchmark}, Conceptual Captions (CC3M) \cite{cc3m}, Conceptual 12M (CC12M) \cite{cc12m} and LAION115M \cite{li2022blip} datasets to pre-train our models. We construct three different pre-training settings, including \textbf{15M}, \textbf{30M}, and \textbf{145M} settings. Each of the settings uses different combinations of pre-training datasets, as shown in Table. The 15M setting is used for the ablation study and to compare our methods with state-of-the-art methods under a fair setup \cite{declip_benchmark}. The 30M and 145M settings are used to evaluate the  scalability of our model.

\begin{table}[h]
\centering
\resizebox{0.8\linewidth}{!}{
\begin{tabular}{@{}ll@{}}
\toprule
Setting & Dataset                             \\ \midrule
15M     & YFCC-15M V2                         \\
30M     & YFCC-15M V2, CC3M, CC12M            \\
145M    & YFCC-15M V2, CC3M, CC12M, LAION115M \\ \bottomrule
\end{tabular}
}
\caption{The used pre-training datasets under different settings.}
\vspace{-10pt}
\label{tbl:setting}
\end{table}


\begin{table*}[t]
\centering
\resizebox{0.7\linewidth}{!}{
\begin{tabular}{@{}lcccccccccc@{}}
\toprule
\multicolumn{1}{l}{} &
  \rotatebox{0}{C10} &
  \rotatebox{0}{C100} &
  \rotatebox{0}{F101} &
  \rotatebox{0}{PETS} &
  \rotatebox{0}{FLOW} &
  \rotatebox{0}{SUN} &
  \rotatebox{0}{DTD} &
  \rotatebox{0}{CAL} &
  \rotatebox{0}{IN} &
  \rotatebox{0}{AVG} \\ \midrule

SLIP~\cite{slip}   & 50.7 & 25.5 & 33.3 & 23.5 & 49.0   & 34.7 & 14.4 & 59.9 & 34.3 & 36.1 \\
MS-CLIP-S~\cite{msclip}  & -    & -    & -    & -    & -    & -    & -    & -    & 36.7 & -    \\
CLIP~\cite{clip}                   & 60.4 & 33.5 & 39.6 & 23.1 & 54.0 & 42.0 & 17.0 & 65.5 & 37.0                 & 41.3                  \\
FILIP~\cite{filip}    & 65.1 & 34.2 & 43.2 & 24.1 & 52.8 & 50.8 & 24   & 68.9 & 39.5 & 44.7 \\
DeCLIP~\cite{declip}    & 72.8 & 40.3 & 49.9 & 36.2 & 60.1 & 48.8 & 26.4 & 72.7 & 43.2 & 50.0 \\
\midrule
CLIP+FDT (Ours)                & 67.7 & 39.9 & 42.9 & 25.8 & 55.5 & 45.5 & 26.5 & 69.6 & 39.3                 & 45.9 \\
DeCLIP+FDT (Ours)               & \textbf{75.7} &
  \textbf{45.2} &
    \textbf{52.9} &
  \textbf{40.7} &
  \textbf{64.6} &
  \textbf{52.0} &
  \textbf{30.7} &
  \textbf{76.2} &
  \textbf{45.8} &
  \textbf{53.8}    \\ \bottomrule
\end{tabular}}

\caption{Zero-shot image classification accuracy (\%) under the 15M setting. The dataset names are abbreviated. 
C10/100 is CIFAR10/100. F101 is Food101. FLOW is Flowers. CAL is Caltech. IN
is ImageNet-1K. ``AVG'' is the average accuracy over all datasets.}
\label{tbl:15m_zs}

\end{table*}

\begin{table*}[ht!]
\centering
\resizebox{0.7\linewidth}{!}{
\begin{tabular}{@{}lccccccccccc@{}}
\toprule
\multicolumn{1}{l}{} &
  \rotatebox{0}{C10} &
  \rotatebox{0}{C100} &
  \rotatebox{0}{F101} &
  \rotatebox{0}{PETS} &
  \rotatebox{0}{FLOW} &
  \rotatebox{0}{SUN} &
  \multicolumn{1}{l}{\rotatebox{0}{CARS}} &
  \rotatebox{0}{DTD} &
  \rotatebox{0}{CAL} &
  \rotatebox{0}{AIR} &
  \rotatebox{0}{AVG} \\ \midrule
SLIP~\cite{slip}  & 87.4 & 69.5 & 71.3 & 70.5 & 91.9 & 66.9 & 27.5 & 65.6 & 86.2 & 27.7 & 66.5 \\
MS-CLIP-S~\cite{msclip}  & 87.2 & 66.7 & 76.0   & 62.1 & 93.8 & 71.7 & 27.5 & 69.4 & 81.6 & \textbf{32.9} & 66.9 \\
CLIP~\cite{clip}                  & 88.3 & 68.6 & 72.1 & 72.5 & 92.6 & 69.5 & 29.8                 & 67.8 & 86.2 & 27.7                     & 67.5                  \\
FILIP~\cite{filip}   & 86.5 & 66.6 & 71.7 & 69.2 & 93   & 69.6 & 30.0   & 66.4 & 85.7 & 27.0   & 66.6 \\
DeCLIP~\cite{declip}   & 89.4 & 69.6 & 75.9 & 71.4 & 95.7 & 71.6 & 30.1 & 66.9 & 89.0   & 26.7 & 68.6 \\
\midrule
CLIP+FDT (Ours)               & 89.1 & \textbf{71.2} & 74.4 & 73.0   & 93.4 & 70.8 & 31.4                 & 69.4 & 87.7 & 27.9                     & 68.8 \\


DeCLIP+FDT (Ours)                & \textbf{89.8}& \textbf{71.2}& \textbf{77.7}& \textbf{73.9}& \textbf{95.7}& \textbf{72.9}& \textbf{33.7}& \textbf{69.6}& \textbf{89.4}& 26.9&  \textbf{70.1}                      \\
\bottomrule
\end{tabular}}
\caption{Linear probing image classification accuracy (\%) under the 15M setting. The dataset names are abbreviated. 
C10/100 is CIFAR10/100. F101 is Food101. FLOW is Flowers. CAL is Caltech. Air is Aircraft. ``AVG'' is the average accuracy over all datasets.}
\label{tbl:15m_lp}
\vspace{-10pt}
\end{table*}

\begin{table*}[ht!]
\centering
\resizebox{0.8\linewidth}{!}{
\begin{tabular}{@{}lcccccccccccc@{}}
\toprule
            & \multicolumn{4}{c}{Flickr30K} & \multicolumn{4}{c}{MSCOCO} & \multicolumn{4}{c}{VQAv2}                       \\
 & \multicolumn{2}{c}{Image Retrieval} & \multicolumn{2}{c}{Text Retrieval} & \multicolumn{2}{c}{Image Retrieval} & \multicolumn{2}{c}{Text Retrieval} &  &  &  &  \\
            & R@1   & R@5   & R@1   & R@5   & R@1   & R@5  & R@1  & R@5  & y/n  & number & other   & overall               \\ \midrule
SLIP~\cite{slip}    & 23.3  & 47.2  & 35.7  & 65.8  & 13.2  & 31.3 & 21.0   & 44.6 & 69.8 & 34.3   & 38.1    & 50.7                  \\
MS-CLIP-S~\cite{msclip}  & -     & -     & -     & -     & 19.4  & 40.8 & 28.5 & 54.1 & -    & -      & -       & -                     \\
CLIP~\cite{clip}    & 27.6  & 53.9  & 42.8  & 71.5  & 15.9  & 36.7 & 24.8 & 49.8 & 67.7 & 31.9   & 33.6    & 47.5                  \\
FILIP~\cite{filip}      & 30.6  & 58.2  & 46.3  & 74.4  & 16.2  & 37.5 & 25.6 & 50.8 & 68.1 & 34.5   & 36.2    & 49.2                  \\
DeCLIP~\cite{declip}      & 35.5  & 63.0  & 51.2  & 80.7  & 19.6  & 41.9 & 30.1 & 55.6 & \textbf{70.3} & 34.9   & 36.9    & 50.4                  \\

\midrule
CLIP+FDT (Ours)  & 32.6  & 58.6  & 51.0  & 78.3  & 19.4  & 40.8 & 29.6 & 55.3 & 67.8 & 34.6   & 39.6    & 50.6 \\
DECLIP+FDT (Ours)  & \textbf{39.4}  & \textbf{66.8}  & \textbf{57.0}  & \textbf{82.3}  & \textbf{22.5}  & \textbf{45.5} & \textbf{34.0} & \textbf{59.6} & 67.8 & \textbf{35.8}   & \textbf{41.3}    & \textbf{51.6} \\ \bottomrule
\end{tabular}
}

\caption{Results of the vision-language tasks under the 15M setting, including the zero-shot image-text retrieval on the Flickr30K and MSCOCO (5K) datasets, and the non-linear probing on VQA v2 dataset.}

\label{tbl:15m_vl}
\vspace{-10pt}
\end{table*}

\noindent{\textbf{Evaluation Protocols.~}}
Following previous work \cite{declip, filip, declip_benchmark}, our method is evaluated on three commonly-used downstream tasks, including zero-shot image classification, linear probe image classification, and zero-shot image-text retrieval. Moreover, we propose a non-linear probe task to evaluate the effectiveness of the learned features for VQA~\cite{vqa}. The FDT-based features are used for all the downstream tasks.

\noindent\emph{Zero-shot image classification.~} In this task, image categories are represented by the text descriptions generated from their names. After extracting the embeddings of these text descriptions and input images by pre-trained encoders, the category of an image can be predicted by choosing the one whose text descriptions have the largest cosine similarity score. 
Following the setting of CLIP and DeCLIP, we construct 80 prompts to evaluate the performance of different approaches. 
We use 9 of the 11 commonly used datasets \cite{declip} for evaluation. The StanfordCars and Aircraft datasets are not used, because the pre-training datasets contain few captions about car models or aircraft types.

\noindent\emph{Linear Probe Image Classification.~} A linear classifier is trained to predict the categories of images based on the FDT-based features of the images. We use 10 of the 11 commonly
used datasets for evaluation. We do not report the results on ImageNet-1K, since conducting hyperparameter sweeping on this dataset is computationally expensive.

\noindent\emph{Image-text retrieval.~}  The image-text retrieval task is evaluated on the Flickr30K \cite{f30k} and MSCOCO \cite{mscoco} dataset. The recalls at different K values (R@K, K = 1, 5, 10) are reported as the evaluation metrics. They are used to measure the percentage of relevant items that match the queries in top-K retrieved items.~We also report rsum, which is obtained by summing all R@K values.

\noindent\emph{Non-linear probe task.~} The task is to evaluate the capability of learned features for vision-language reasoning tasks. The FDT-based embeddings of an image and its questions are concatenated and fed to two fully-connected layers with non-linear activation to predict the answer. More details can be found in the supplementary materials.

\begin{table*}[t]
\centering
\resizebox{0.9\linewidth}{!}{
\centering
\begin{tabular}{@{}lccc|ccc|cccc@{}}
\toprule
 &
   &
  ZS CLS &
  LP CLS &
  \multicolumn{3}{c|}{ZS-Flickr30K} &
   &
  ZS-MSCOCO &
   &
  VQAv2 \\
 &
  Setting &
  AVG Acc &
  AVG Acc &
  IR R@1 &
  TR R@1 &
  rsum &
  IR R@1 &
  TR R@1 &
  rsum &
  overall \\ \midrule
CLIP &
  15M &
  41.3 &
  67.5 &
  27.6 &
  42.8 &
  343.1 &
  15.9 &
  24.8 &
  236.8 &
  47.5 \\
CLIP+FDT &
  15M &
  45.9($\uparrow$4.6) &
  68.8($\uparrow$1.3) &
  32.6($\uparrow$5.0) &
  51.0($\uparrow$8.2) &
  376.5($\uparrow$33.4) &
  19.4($\uparrow$3.5) &
  29.6($\uparrow$4.8) &
  263.1($\uparrow$26.3) &
  50.6($\uparrow$3.1) \\ \midrule
CLIP &
  30M &
  56.8 &
  73.8 &
  43.6 &
  58.8 &
  431.3 &
  23.3 &
  34.8 &
  300.8 &
  50.6 \\
CLIP+FDT &
  30M &
  61.2($\uparrow$ 4.4) &
  75.6 ($\uparrow$ 1.8) &
  52.5($\uparrow$8.9) &
  70.8($\uparrow$12.0) &
  474.2($\uparrow$42.9) &
  28.3($\uparrow$5.0) &
  43($\uparrow$8.2) &
  337.1 ($\uparrow$36.3) &
  53.4($\uparrow$2.8) \\ \midrule
CLIP &
  145M &
  64 &
  82.1 &
  52.6 &
  67.9 &
  469.8 &
  29.3 &
  42.1 &
  335.2 &
  53.1 \\
CLIP+FDT &
  145M &
  69.0($\uparrow$ 5.0) &
  82.3 ($\uparrow$ 0.2) &
  56.3($\uparrow$3.7) &
  75.9($\uparrow$8.0) &
  489.4($\uparrow$19.6) &
  31.0($\uparrow$1.7) &
  46.4($\uparrow$4.3) &
  353.0($\uparrow$17.8) &
  55.2($\uparrow$2.1) \\ \bottomrule
\end{tabular}


}
\caption{Ablation study results when using different scales of training data. ``ZS'' means zero-shot. ``AVG'' is average.  ``ACC'' is accuracy. ``LP'' stands for linear prob.  ``CLS'' represents classification. ``IR'' and ``TR'' are image retrieval and text retrieval, respectively.}
\label{tbl:data_scale}
\vspace{-5pt}
\end{table*}

\begin{table*}[ht!]
\centering
\resizebox{0.9\linewidth}{!}{

\begin{tabular}{lcc|ccc|cccc}
\toprule
 &
  ZS CLS &
  LP CLS &
  \multicolumn{3}{c|}{ZS-Flickr30K} &
   &
  ZS-MSCOCO &
   &
  VQAv2 \\
 &
  AVG Acc &
  AVG Acc &
  IR R@1 &
  TR R@1 &
  rsum &
  IR R@1 &
  TR R@1 &
  rsum &
  Overall \\ \midrule
CLIP-ViT-B/32 &
  41.3 &
  67.5 &
  27.6 &
  42.8 &
  343.1 &
  15.9 &
  24.8 &
  236.8 &
  47.5 \\
CLIP-ViT-B/32+FDT &
  45.9($\uparrow$4.6) &
  68.8($\uparrow$1.3) &
  32.6($\uparrow$5.0) &
  51.0($\uparrow$8.2) &
  376.5($\uparrow$33.4) &
  19.4($\uparrow$3.5) &
  29.6($\uparrow$4.8) &
  263.1($\uparrow$26.3) &
  50.6($\uparrow$3.1) \\ \midrule
CLIP-ViT-B/16 &
  45.2 &
  68.8 &
  35.3 &
  50.5 &
  387.8 &
  19.3 &
  29.7 &
  263.6 &
  49.2 \\
CLIP-ViT-B/16+FDT &
  49.9($\uparrow$4.7) &
  71.3($\uparrow$2.5) &
  41.6($\uparrow$6.3) &
  60.8($\uparrow$10.3) &
  425.5($\uparrow$37.7) &
  23.4($\uparrow$4.1) &
  35.3($\uparrow$5.6) &
  295.4 ($\uparrow$31.8) &
  54.3($\uparrow$5.1) \\ \midrule
CLIP-Swin-B &
  39.6 &
  68.5 &
  30.5 &
  48.5 &
  368.1 &
  17.7 &
  26.0 &
  247.6 &
  46.5 \\
CLIP-Swin-B+FDT &
  42.4($\uparrow$2.8) &
  70.7($\uparrow$2.2) &
  39.6($\uparrow$9.1) &
  57.9($\uparrow$9.4) &
  415.5($\uparrow$47.4) &
  22.3($\uparrow$4.6) &
  33.8($\uparrow$7.8) &
  288.3($\uparrow$40.7) &
  51.6($\uparrow$5.1) \\ \bottomrule
\end{tabular}

}
\caption{Ablation Study results when using different image encoder architectures. ``ZS'' means zero-shot. ``AVG'' is average.  ``ACC'' is accuracy. ``LP'' stands for linear prob.  ``CLS'' represents classification. ``IR'' and ``TR'' are image retrieval and text retrieval.}
\label{tbl:enc_abl}
\vspace{-10pt}
\end{table*}

\noindent{\textbf{Implementation Details.~}} We evaluate our method by incorporating it into two state-of-the-art contrastive vision-language pre-training approaches, namely CLIP \cite{clip} and DECLIP \cite{declip}. Our implementation is based on the open-source PyTorch implementation\footnote{https://github.com/Sense-GVT/DeCLIP} of the two methods. We use 16384 tokens, each with 512 dimensions. Please refer to the supplementary material for detailed information.

\begin{table*}[t]
\centering
\resizebox{0.75\linewidth}{!}{
\begin{tabular}{@{}lcc|ccc|cccc@{}}
\toprule
\multicolumn{1}{l}{} & ZS CLS  & LP CLS  & \multicolumn{3}{c|}{ZS-Flickr30K} &        & ZS-MSCOCO &       & VQAv2   \\
FDT size           & AVG Acc & AVG Acc & IR R@1     & TR R@1    & rsum     & IR R@1 & TR R@1    & rsum  & overall \\ \midrule
-                    & 41.3    & 67.5    & 27.6       & 42.8      & 343.1    & 15.9   & 24.8      & 236.8 & 47.5    \\
8192                 & 42.8    & 67.9    & 32.7       & 50.6      & 374.6    & 18.5   & 29.1      & 258.1 & 50.1    \\
16384                & \textbf{45.9}    & \textbf{68.8}    & 32.6       & \textbf{51.0}        & 376.5    & \textbf{19.4}   & 29.6      & \textbf{263.1} & 50.6    \\
24576                & 45.2    &68.6         & \textbf{33.3}       & 50.4      & \textbf{378.5}    & 18.6   & \textbf{29.7}      & \textbf{263.1} & \textbf{51.4}    \\ \bottomrule
\end{tabular}}
\caption{Results of the models with different FDT sizes. The row whose FDT value is ``-'' represents the original CLIP model. ``ZS'' means zero-shot. ``AVG'' is average.  ``ACC'' is accuracy. ``LP'' stands for linear prob.  ``CLS'' represents classification. ``IR'' and ``TR'' are image retrieval and text retrieval.}
\label{tbl:fdt_size}
\vspace{-5pt}
\end{table*}

\begin{table*}[ht!]
\centering
\resizebox{0.75\linewidth}{!}{
\begin{tabular}{@{}lcc|ccc|ccc|c@{}}
\toprule
                     & ZS CLS  & LP CLS  & \multicolumn{3}{c|}{ZS-Flickr30K} &        & ZS-MSCOCO &       & VQAv2   \\
                     & AVG Acc & AVG Acc & IR R@1     & TR R@1    & rsum     & IR R@1 & TR R@1    & rsum  & overall \\ \midrule
CLIP                 & 41.3    & 67.5    & 27.6       & 42.8      & 343.1    & 15.9   & 24.8      & 236.8 & 47.5    \\
CLIP+FDT$_{\mathrm{Softmax}}$ *   & 5.2     & -       &5.4            &1.7           &45.5          &2.4        &0.8           &26.2       &-         \\
CLIP+FDT$_{\mathrm{Sparsemax}}$ * & 32.4    & -       &10.5            &32.5           &242.4          & 6.0        & 18.3           & 157.5       &-      \\ \midrule
CLIP+FDT$_{\mathrm{Softmax}}$     & 43.9    &68.7         & 33.3       & 47.9      & 377.6    & 19.2   & 28.3      & 258.8 & 47.9    \\
CLIP+FDT$_{\mathrm{Sparsemax}}$   & 45.9    & 68.8    & 32.6       & 51.0      & 376.5    & 19.4   & 29.6      & 263.1 & 50.6    \\ \bottomrule
\end{tabular}
}
\caption{Results of models trained with (Sparsemax) and without (Softmax) sparse constraints. The rows marked with ``*'' are the results when using FDT weights as features (see Section \ref{sec_exp_sparse}). ``ZS'' means zero-shot. ``AVG'' is average.  ``ACC'' is accuracy. ``LP'' stands for linear prob.  ``CLS'' represents classification. ``IR'' and ``TR'' are image retrieval and text retrieval.} 
\label{tbl:sparse}
\vspace{-5pt}
\end{table*}

\subsection{Comparison with State-of-the-Art Approaches}
We compare our method with the state-of-the-art CLIP family approaches on the benchmark proposed in \cite{declip_benchmark}. In this benchmark, methods are compared fairly by pre-training them using the same training recipe and data (our 15M setting).~Note that the original paper only reports the results for zero-shot classification on the ImageNet dataset, and the results of other tasks are obtained by directly applying the released checkpoints for evaluation.

The results for zero-shot image classification, linear prob image classification, and vision-language reasoning tasks are reported in Table \ref{tbl:15m_zs}, \ref{tbl:15m_lp}, and \ref{tbl:15m_vl}, respectively. 
First, we observe that using the proposed FDT-based representation with CLIP (i.e., CLIP+FDT) can achieve significant performance improvement over CLIP on all the downstream tasks. 
Notably, CLIP+FDT can outperform FILIP \cite{filip}, which aligns image and text information at the fine-grained patch and language token levels. The results suggest that aligning global cross-modal information in a unified space is more effective than directly aligning fine-grained patches and language tokens with different granularities.
Interestingly, the linear probe results show that CLIP+FDT can learn a comparable image encoder with DeCLIP, which applies various self-supervised pretext tasks that have already been proven effective for visual recognition. One possible reason is that aligning the information in a unified space helps our model better leverage semantic supervision signals in the language domain. We can also see that our method can significantly improve DeCLIP for all the tasks and achieve state-of-the-art performance on the benchmark. It shows that our approach is compatible with self-supervised learning tasks to improve CLIP. Moreover, FDT can improve the VQAv2 task, which requires the capability of collaborative multi-modal reasoning and content understanding.

\subsection{Ablation Study}
In this section, we conduct ablation studies to investigate how different factors influence the performance of our approach. These factors include the pre-training data scale, image encoder architecture, and several design choices of our method. Throughout the ablation study, we use the CLIP model as the baseline to save computation costs.

\noindent{\textbf{Pretraining Data Scale.}} We evaluate the performance of our methods on different pre-training data scales by further pre-training the model on 30M and 145M data. According to the results presented in Table \ref{tbl:data_scale}, our method still achieves improved performance for all the downstream tasks when pre-trained on larger datasets. We also note that the improvement for the linear probing setting is minor when pre-trained on 145M data. We assume this is because the performance of the model saturates. To further improve the performance of the image encoder, a more vision-specific training task is needed. Note that using FDT still achieves significant performance improvements on 145M data for other tasks. Interestingly, our model achieves significant improvements on the 30M data. One possible reason is that our FDT can benefit significantly from cleaning supervision information in the CC3M \cite{cc3m} and CC12M \cite{cc12m} datasets. We have similar observations for the VQAv2 task.

\noindent{\textbf{Image Encoder Architecture.~}} We evaluate the influence of different image encoder architectures on our proposed method, and the results are reported in Table \ref{tbl:enc_abl}. We observe that our method still significantly outperforms CLIP when using different types of image encoders. Additionally, FDT slightly adds an average of 6\% more parameters, 13\% more training time, and 12\% less throughput when using different encoder architectures. The detailed results can be found in the supplementary materials.



\noindent{\textbf{FDT Numbers.}} The performance of models trained with different learnable token numbers are shown in Table \ref{tbl:fdt_size}. We can see that using 8192 tokens can already achieve an improvement over CLIP. Increasing the FDT size to 16384 obtains a more significant improvement than 8192, since it can encode more types of information. Furthermore, growing the FDT size to 24576 achieves a slight improvement over 16384 for the zero-shot image-text retrieval task on the Flickr30K dataset and VQA task. We set the FDT size as 16384 in our implementation because it achieves the best performance-efficiency tradeoff.

\noindent{\textbf{Sparse Constraints.}} 
\label{sec_exp_sparse}
In this section, we aim to demonstrate that applying sparse constraints helps the model learn better cross-modal correspondence, where the same cross-modal information is represented using the same subset of FDT. To this end, we evaluate the performance when using the FDT weights (Equation \ref{eq:img_softmax}, \ref{eq:text_softmax} and \ref{eq:sparsemax}) of each image or sentence as the features for zero-shot image classification and image-text retrieval tasks. The results are reported in Table~\ref{tbl:sparse}. From the table, we can see that using sparse constraints (Sparsemax) achieves significantly better performance for all tasks. The results demonstrate that adding sparse constraints to FDT weights can lead to better cross-modal correspondence. 
Additionally, we can also see that without sparse constraints (Softmax), FDT-based features can also achieve significant performance over CLIP. Adding a sparse constraint (Sparsemax) achieves a larger performance improvement. This is because the granularities are further unified by representing the same cross-modal information with the same token set.




\begin{figure}[t!]
\begin{center}
    \includegraphics[width=\linewidth]{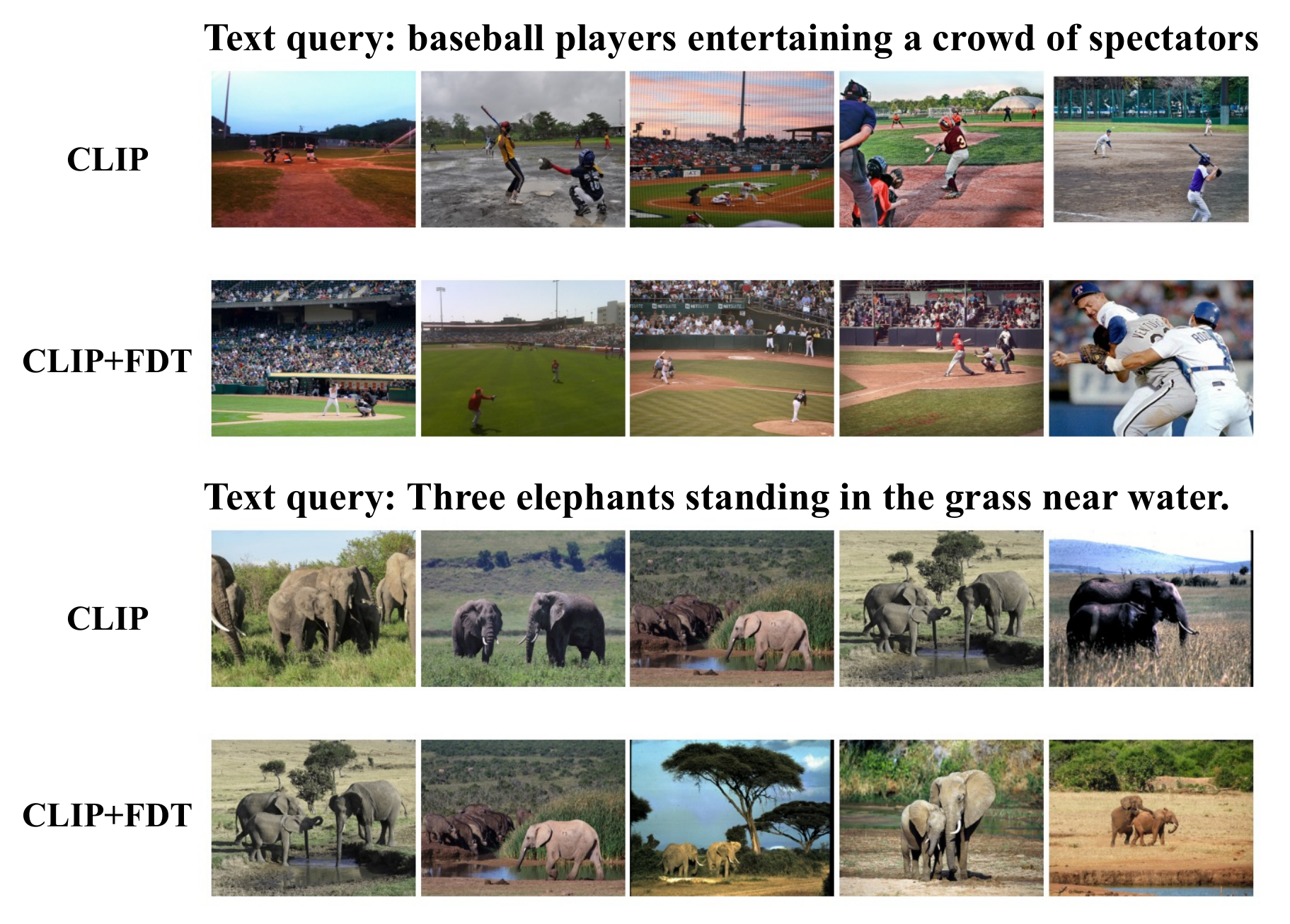}
\end{center}
\vspace{-10pt}
\caption{Examples shows the top-5 retrieved images for the given text queries for the text-to-image retrieval task on MSCOCO.}
\label{fig:coco_case}
\vspace{-15pt}
\end{figure}

\begin{figure*}[ht!]
\begin{center}
    \includegraphics[width=0.95\linewidth] {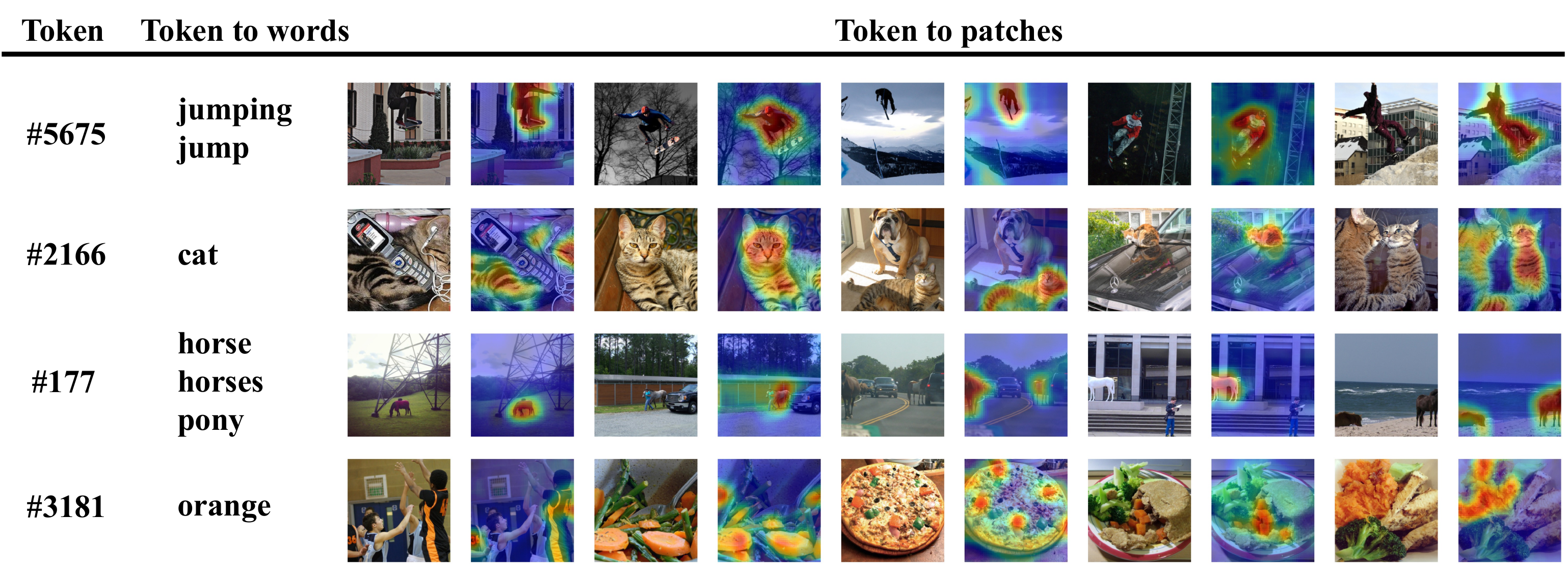}
\end{center}
\vspace{-5pt}
\caption{Example of the top-5 most relevant image patches and text tokens of four FDT tokens. Note that the redundant text tokens in the top-5 are removed. The color of the heatmap from blue to red denotes the relevance between patches and FDT from small to large.}
\label{fig:code_vis}
\vspace{-10pt}
\end{figure*}

\subsection{Analysis of the Completeness of Alignment}

Since the granularities of image and text information are inconsistent, the learned model may fail to capture key semantic concepts \cite{learning_concepts}. In this experiment, we empirically evaluate whether unifying the granularities through the proposed FDT can alleviate the problem.  The model pretrained on the 145M dataset is used for this evaluation.

To this end, we design a probing experiment on the MSCOCO dataset. Using the object detection annotations in the training split of MSCOCO, we construct 305,723 sentence pairs. For each sentence pair, one \emph{matched sentence} describes all objects in an image, while the other \emph{partially matched sentence} only captures part of the objects. Please refer to the supplementary material for more details about how we constructed these sentence pairs. 

We then use pre-trained models to extract the embeddings of images and sentences and compute the similarity scores between the images and these constructed sentences. If the learned model comprehensively captures the semantic concepts, the similarity between an image and its matched sentence should be higher than that between the partially matched sentence. We found that the CLIP+FDT models can meet our expectation in 68.2 \% of all sentence pairs, surpassing the CLIP model by 7.6\%. The results demonstrate that FDT can help the CLIP model more comprehensively capture various semantic concepts. We assume that this is because the FDT serve as the prior knowledge that guides encoders to extract cross-modally shared high-level semantic concepts. This not only facilitates cross-modal interactions but also helps encoders capture semantic information from images and texts more comprehensively.

In addition, we show two cases for the text-to-image retrieval task in Figure \ref{fig:coco_case}. We can see that the images retrieved by CLIP ignore some important concepts described in the text queries. For example, in terms of the text query ``baseball players entertaining a crowd of spectators'', four out of the five images retrieved by the CLIP models contain baseball players only but with no spectators. Moreover, the image containing spectators is ranked lower than the two images without spectators. In contrast, FDT can retrieve images that contain both baseball players and spectators. More results are provided in the supplementary material.



\subsection{Visualization of Learned FDT}
To explicitly show the cross-modal correspondence learned by our FDT, we visualize the top-5 most relevant image patches and text tokens (using Equation \ref{eq:img_relevance} and \ref{eq:text_relevance}) of four FDT tokens in Figure~\ref{fig:code_vis}. The MSCOCO dataset and the model pretrained on the 145M dataset are used for visualization. The example cases show that each token captures different types of cross-modal correspondence, including actions (jump/jumping), objects, and attributes (orange color). Moreover, the learned FDT can potentially detect correspondent patches from the images. For example, the second token has high relevance values with patches of cats, while having low relevance with other patches. More results can be found in the supplementary material.

\section{Conclusions}
In this paper, we introduce a new multimodal presentation using finite discrete tokens (FDT). Specifically, a set of learnable tokens shared by all modalities are used to represent multimodal information conveyed in the image and text modalities. 
Our approach is a light-weighted way of fulfilling cross-modal interaction, where FDT serves as multimodal anchors to capture information from each input with better completeness. This help alleviate the model degradation problem commonly observed in vanilla CLIP models.~Our FDT can be trained with the contrastive learning scheme from scratch without cold-start problems.~Both quantitative and qualitative results demonstrate that FDT representations achieve better cross-modal alignment and performance on various downstream tasks, including image classification, cross-modal retrieval, and VQA. Additionally, the learned FDT capture meaningful cross-modal correspondence, ranging
from objects to actions and attributes.



%% file: sections/supp_final.tex
\section{Pre-training Implementation Details}
We implement the projecting function that maps patch or language token features to the FDT space as a fully-connected layer with GELU activation (see Section 3.2). Two different projecting functions are applied for mapping patch and language token features, respectively. We regularize the FDT using weight decay, with a rate of 0.1. We set the batch sizes as 4096, 8192, and 32768 when pretraining the models under the 15M, 30M, and 145M settings, respectively. To ensure a fair comparison with the DECLIP~\cite{declip} and FILIP~\cite{filip} models, we use the same data augmentation as these models when training the CLIP and CLIP+FDT models. Consequently, our reported results of the CLIP model on the 15M setting are better than those reported in the 15M benchmark~\cite{declip_benchmark}. We train ViT-B/32 based \cite{vit} models considering our limited computation resource.  The input image resolution is 224 $\times$ 224, and the maximal input language token number is 77. Following \cite{declip_benchmark}, we apply the AdamW optimizer \cite{loshchilov2018decoupled} with a weight decay rate of 0.1 during pre-training. The learning rate is first linearly increased to 0.001 with one epoch for warmup, and then decayed to 0 following the cosine strategy \cite{cosinedecay}.~We use NVIDIA A100 GPUs for pre-training.

\section{Downstream Implementation Details}
\subsection{Downstream Datasets}
\noindent{\textbf{Image Classification Tasks.~}}Following \cite{declip}, we evaluate our method on 11 datasets, including CIFAR-10~\cite{cifar10}, CIFAR-100~\cite{cifar10}, SUN397~\cite{sun397}, Stanford Cars~\cite{cars}, FGVC Aircraft~\cite{aircraft},
Describable Textures~\cite{dtd}, Oxford-IIIT Pets~\cite{pets}, Caltech-101~\cite{caltech}, Oxford Flowers 102~\cite{flower}, Food-101~\cite{food101}, and ImageNet-1K~\cite{imagenet}. 

\noindent{\textbf{Image-Text Retrieval.~}}Our method is tested on two standard benchmarks: Flickr30K~\cite{f30k} and MSCOCO~\cite{mscoco}. For MSCOCO, we report the results on the 5K setting.

\noindent{\textbf{Non-Linear Probe task.~}}We conduct the experiments on the VQAv2 dataset \cite{vqa}. Following the standard protocol \cite{meter}, we train the models with both training and validation data, and test the models on the test-dev set.

\subsection{Implementation Details}
\noindent{\textbf{Zero-shot Image Classification.~}}For a fair comparison, we use the domain-specific prompts and category names proposed by CLIP~\cite{clip}. Note that we do not report the results on the StanfordCars and Aircraft datasets, because the pertaining datasets contain few captions about the category names of these datasets. For example, only 0.04\% and 0\% of descriptions contain aircraft and car category names on the 15M setting.

\noindent{\textbf{Linear Probe Image Classification.~}}We train a logistic regression classifier using L-BFGS, following CLIP \cite{clip}. We set the maximum iterations number to 1,000, and determine the L2 regularization weights following DECLIP's hyperparameter sweeping strategy \cite{declip}. We do not report the results on the ImageNet-1K dataset, due to the high computational cost of conducting hyperparameter sweeping on the dataset.

\noindent{\textbf{Non-linear Probe Task.~}}The downstream task head consists of a fully-connected layer with GELU activation and a fully-connected layer. The extracted FDT features of images and questions are concatenated and then fed to the downstream task head to predict the answers. The encoders and FDT are frozen during the training. The downstream head is optimized by the AdamW optimizer \cite{loshchilov2018decoupled}. We set the learning rate as 0.005, and decay it to 0 following the cosine strategy \cite{cosinedecay}.

\section{Completeness Probing Experiment Details}
Given an image that contains $N$ objects, its \textit{matched sentence}
is ``An photo contains $o_1$, $o_2$ ..., $o_{N-1}$, and $o_N$'', where $o_i$ is the name of the $i$-th object in the images and all the objects are included. For the \textit{partially matched sentence}, we randomly remove an object and use the remaining $N-1$ objects to construct a caption. For example, if the $N$-th object is removed, the partially matched sentence is ``An photo contains $o_1$, $o_2$ ..., and $o_{N-1}$''. We can construct $N$ partially matched sentences for the image, resulting in $N$ \textit{sentence pairs} for the image. In our experiments, we obtain the object presence information of images based on the object detection annotations of the MSCOCO \cite{mscoco} dataset. We construct 305,723 sentence pairs using all images in the MSCOCO training split.

\section{FDT Visualization Details}

We use the model pre-trained on the 145M setting for visualization because it achieves the best performance. To visualize an FDT token, we first calculate its relevance score between patches/language tokens following Equations 4 and 6 without using max-pooing. We then display the relevance scores between the FDT token and the images corresponding to the top-5 most relevant patches, since we find that the patches alone cannot fully convey the object information. We increase the resolution by reducing the patch stride to 4, following the method proposed in \cite{amir2021deep}. For text modality, we show the top-5 most relevant language tokens of the FDT token.

\section{Additional Experiment Results}

\subsection{Text-to-Image Retrieval Cases}

\begin{figure}[t!]
\begin{center}
    \includegraphics[width=1\linewidth]{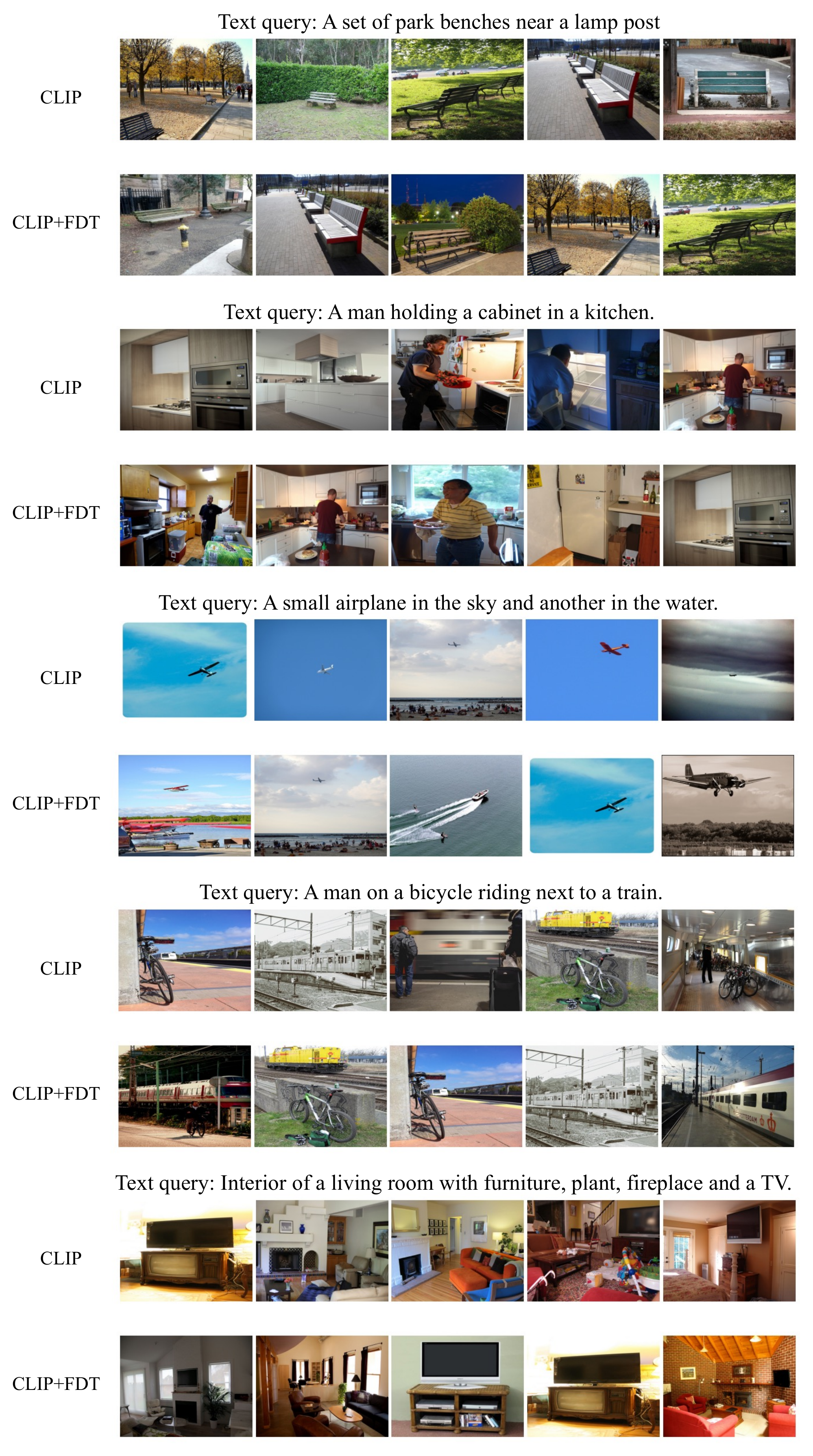}
\end{center}
\vspace{-5pt}
\caption{Examples show the top-5 retrieved images for the given text queries in the text-to-image retrieval task on MSCOCO.}
\label{fig:coco_case_supp}
\end{figure}

We further provide five cases for the text-to-image retrieval task in Figure \ref{fig:coco_case_supp}. We have the same observation that the images retrieved by the CLIP+FDT well match the text queries, while those retrieved by the CLIP models often overlook important concepts mentioned in the text queries.

\vspace{-5pt}

\subsection{Visualization of Learned FDT}

\begin{figure*}[t]
\begin{center}
    \includegraphics[width=0.8\linewidth]{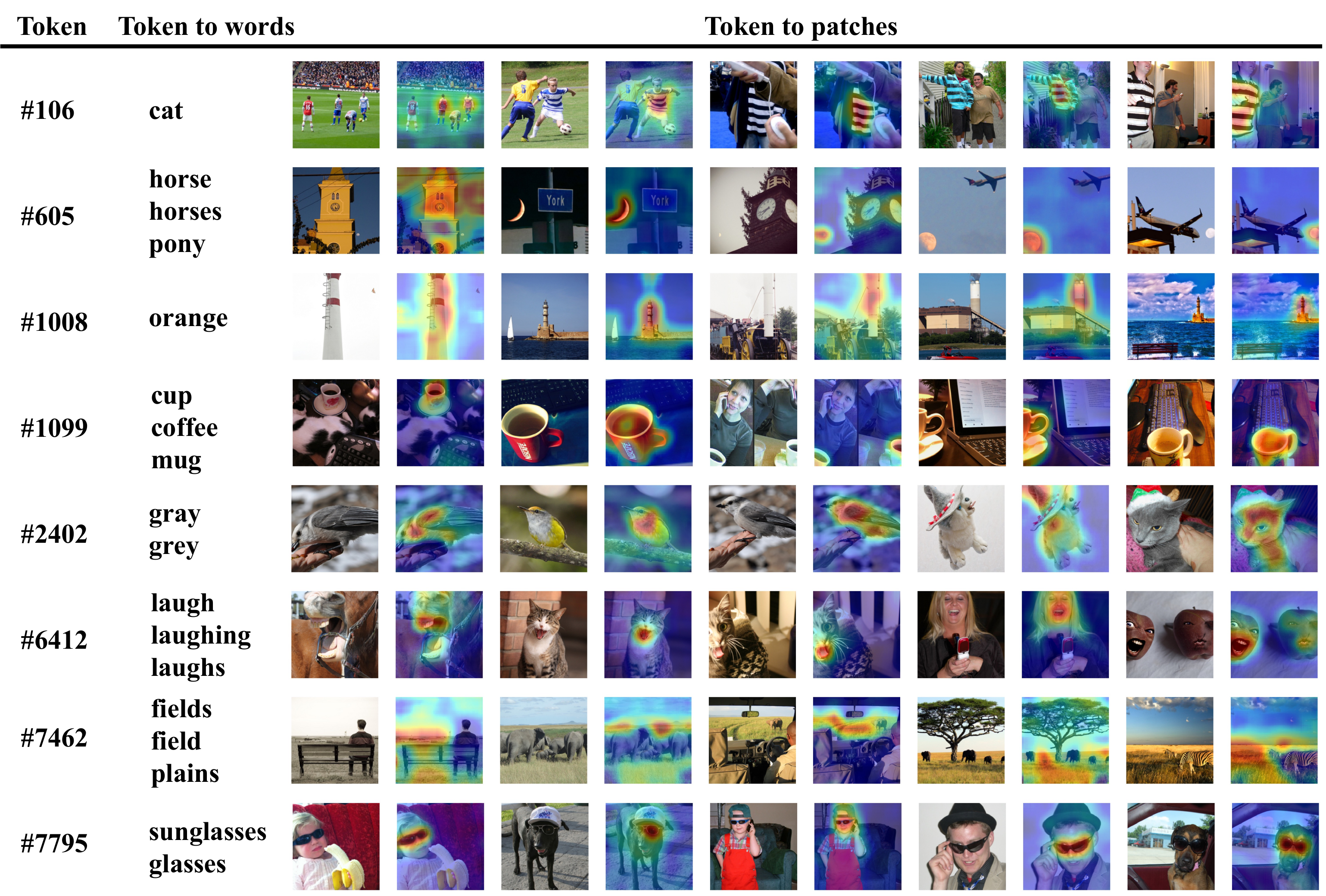}
\end{center}
\vspace{-5pt}
\caption{The top-5 most relevant image patches and text tokens of eight FDT tokens. Note that the redundant text tokens in the top-5 are removed. The color of the heatmap from blue to red denotes the relevance between patches and FDT from small to large.}
\label{fig:fdt}
\end{figure*}

We present eight learned FDT in Figure \ref{fig:fdt}. These cases further show that FDT can learn meaningful cross-modal correspondence. 

\vspace{-5pt}

\subsection{Pretraining Data Scale}

The results of the models pretrained with different scales of training data are reported in Table \ref{tbl:scale_zs_cls}, \ref{tbl:scale_lp_cls}, \ref{tbl:scale_zs_itr}, and \ref{tbl:scale_vqa}.

\begin{table*}[t!]

\centering
\resizebox{0.6\linewidth}{!}{
\begin{tabular}{@{}ccccccccccc@{}}
\toprule
\multicolumn{1}{l}{} &
  \rotatebox{0}{C10} &
  \rotatebox{0}{C100} &
  \rotatebox{0}{F101} &
  \rotatebox{0}{PETS} &
  \rotatebox{0}{FLOWE} &
  \rotatebox{0}{SUN} &
  \rotatebox{0}{DTD} &
  \rotatebox{0}{CAL} &
  \rotatebox{0}{IN} &
  \rotatebox{0}{AVG} \\ \midrule
\multicolumn{1}{l}{15M}  &      &      &      &      &      &      &      &      &                      &                       \\ \midrule
CLIP                     & 60.4 & 33.5 & 39.6 & 23.1 & 54.0 & 42.0 & 17.0 & 65.5 & 37.0                 & 41.3                  \\
CLIP+FDT                & 67.7 & 39.9 & 42.9 & 25.8 & 55.5 & 45.5 & 26.5 & 69.6 & 39.3                 & 45.9 ($\uparrow$ 4.6) \\ \midrule
\multicolumn{1}{l}{30M}  &      &      &      &      &      &      &      &      & \multicolumn{1}{l}{} & \multicolumn{1}{l}{}  \\ \midrule
CLIP                     & 77.2 & 48.1 & 59.1 & 58.4 & 58.2 & 52.6 & 28.0 & 80.8 & 48.8                 & 56.8                  \\
CLIP+FDT                & 81.9 & 56.5 & 62.6 & 62.3 & 59.5 & 56.7 & 33.6 & 84.8 & 53.3                 & 61.2($\uparrow$ 4.4)  \\ \midrule
\multicolumn{1}{l}{145M} &      &      &      &      &      &      &      &      & \multicolumn{1}{l}{} & \multicolumn{1}{l}{}  \\ \midrule
CLIP                     & 80.9 & 53.9 & 69.1 & 68.9 & 59.3 & 52.1 & 43.0 & 90.1 & 59.0                 & 64.0                  \\
CLIP+FDT                & 87.1 & 63.7 & 73.5 & 77.0 & 65.0 & 56.2 & 47.7 & 90.5 & 60.4                 & 69.0($\uparrow$ 5.0)    \\ \bottomrule
\end{tabular}}
\vspace{-5pt}
\caption{Zero-shot image classification accuracy (\%) when using different scales of training data. The dataset names are abbreviated. C10/100 is CIFAR10/100. F101 is Food101. FLOW is Flowers. CAL is Caltech. IN is ImageNet-1K. ``AVG'' is the average accuracy over all datasets.}
\label{tbl:scale_zs_cls}
\end{table*}

\begin{table*}[t!]
\centering
\resizebox{0.65\linewidth}{!}{
\begin{tabular}{@{}cccccccccccc@{}}
\toprule
\multicolumn{1}{l}{} &
  \rotatebox{0}{C10} &
  \rotatebox{0}{C100} &
  \rotatebox{0}{F101} &
  \rotatebox{0}{PETS} &
  \rotatebox{0}{FLOW} &
  \rotatebox{0}{SUN} &
  \multicolumn{1}{l}{\rotatebox{0}{CARS}} &
  \rotatebox{0}{DTD} &
  \rotatebox{0}{CAL} &
  \rotatebox{0}{AIR} &
  \rotatebox{0}{AVG} \\ \midrule
\multicolumn{1}{l}{15M}  &      &      &      &      &      &      & \multicolumn{1}{l}{} &      &      &                          &                       \\ \midrule
CLIP                     & 88.3 & 68.6 & 72.1 & 72.5 & 92.6 & 69.5 & 29.8                 & 67.8 & 86.2 & 27.7                     & 67.5                  \\
CLIP+FDT                & 89.1 & 71.2 & 74.4 & 73.0   & 93.4 & 70.8 & 31.4                 & 69.4 & 87.7 & 27.9                     & 68.8 ($\uparrow$ 1.3) \\ \midrule
\multicolumn{1}{l}{30M}  &      &      &      &      &      &      & \multicolumn{1}{l}{} &      &      & \multicolumn{1}{l}{}     & \multicolumn{1}{l}{}  \\ \midrule
CLIP                     & 92.0   & 74.7 & 78.8 & 80.7 & 93.7 & 72.6 & 55.9                 & 71.4 & 88.6 & 29.7                     & 73.8                  \\
CLIP+FDT                & 93.8 & 77.8 & 81.6 & 82.6 & 94.5 & 74.3 & 54.4                 & 73.9 & 92.3 & 30.9                     & 75.6 ($\uparrow$ 1.8) \\ \midrule
\multicolumn{1}{l}{145M} &      &      &      &      &      &      & \multicolumn{1}{l}{} &      &      & \multicolumn{1}{l}{}     & \multicolumn{1}{l}{}  \\ \midrule
CLIP                & 95.2 & 80.6 & 86.1 & 87.5 & 96.5 & 76.3 & 87.6                 & 77.2 & 94.7 & 39.5  & 82.1                      \\
CLIP+FDT                     & 94.8 & 80.8 & 85.5 & 85.8 & 95.7 & 75.9 & 88.1                 & 78.5 & 94.6 & 42.9                     & 82.3 ($\uparrow$ 0.2)                 \\\bottomrule
\end{tabular}}
\vspace{-5pt}
\caption{Linear probing image classification accuracy (\%) when using different scales of training data. The dataset names are abbreviated. C10/100 is CIFAR10/100. F101 is Food101. FLOW is Flowers. CAL is Caltech. Air is Aircraft. ``AVG'' is the average accuracy over all datasets.}
\label{tbl:scale_lp_cls}
\end{table*}

\begin{table*}[t!]
\centering
\resizebox{0.85\linewidth}{!}{
\begin{tabular}{@{}lcccccccccccccc@{}}
\toprule
             & \multicolumn{7}{c}{Flickr30K}                   & \multicolumn{7}{c}{MSCOCO}                       \\ \midrule
 &
  \multicolumn{3}{c}{Image Retrieval} &
  \multicolumn{3}{c}{Text Retrieval} &
   &
  \multicolumn{3}{c}{Image Retrieval} &
  \multicolumn{3}{c}{Text Retrieval} &
   \\
             & R@1  & R@5  & R@10 & R@1  & R@5  & R@10 & rsum  & R@1  & R@5  & R@10 & R@1  & R@5  & R@10  & rsum  \\ \midrule
15M setting  &      &      &      &      &      &      &       &      &      &      &      &      &       &       \\ \midrule
CLIP         & 27.6 & 53.9 & 64.4 & 42.8 & 71.5 & 82.9 & 343.1 & 15.9 & 36.7 & 47.8 & 24.8 & 49.8 & 61.8  & 236.8 \\
CLIP + FDT  & 32.6 & 58.6 & 68.5 & 51.0 & 78.3 & 87.5 & 376.5 ($\uparrow$ 33.4) & 19.4 & 40.8 & 51.9 & 29.6 & 55.3 & 66.1  & 263.1 ($\uparrow$ 26.3) \\ \midrule
30M setting  &      &      &      &      &      &      &       &      &      &      &      &      &       &       \\ \midrule
CLIP         & 43.6 & 72.8 & 81.3 & 58.8 & 84.2 & 90.6 & 431.3 & 23.3 & 46.9 & 58.6 & 34.8 & 63.3 & 73.9  & 300.8 \\
CLIP + FDT  & 52.5 & 78.7 & 86.4 & 70.8 & 90.8 & 95.0 & 474.2 ($\uparrow$ 42.9) & 28.3 & 53.3 & 64.3 & 43.0 & 69.0 & 79.2  & 337.1 ($\uparrow$ 36.3)\\ \midrule
145M setting &      &      &      &      &      &      &       &      &      &      &      &      &       &       \\ \midrule
CLIP         & 52.6 & 78.5 & 86.4 & 67.9 & 89.9 & 94.5 & 469.8 & 29.3 & 54.1 & 65.4 & 42.1 & 67.1 & 77.2 & 335.2 \\
CLIP + FDT &
  56.3 &
  80.7 &
  87.6 &
  75.9 &
  93.6 &
  95.3 &
  489.4 ($\uparrow$ 19.6) &
  31.0 &
  55.7 &
  66.7 &
  46.4 &
  71.9 &
  81.3 &
  353.0 ($\uparrow$ 17.8) \\ \bottomrule
\end{tabular}}
\vspace{-5pt}
\caption{Zero-shot image-text retrieval results on the Flickr30K and MSCOCO (5K) datasets when using different scales of training data.}
\label{tbl:scale_zs_itr}
\end{table*}

\begin{table*}[t]
\centering
\resizebox{0.32\linewidth}{!}{
\begin{tabular}{@{}lcccc@{}}
\toprule
             & y/n  & number & other & overall               \\ \midrule
15M setting  &      &        &       &                       \\ \midrule
CLIP         & 67.7 & 31.9   & 33.6  & 47.5                  \\
CLIP + FDT  & 67.8 & 34.6   & 39.6  & 50.6 ($\uparrow$ 3.1) \\ \midrule
30M setting  &      &        &       &                       \\ \midrule
CLIP         & 69.7 & 34.8   & 37.8  & 50.6                  \\
CLIP + FDT  & 68.8 & 36.4   & 42.0  & 53.4 ($\uparrow$ 2.8) \\ \midrule
145M setting &      &        &       &                       \\ \midrule
CLIP         & 70.9 & 36.5   & 41.7  & 53.1                  \\
CLIP + FDT  & 71.5 & 37.9   & 45.2  & 55.2 ($\uparrow$ 2.1) \\ \bottomrule
\end{tabular}}
\vspace{-5pt}
\caption{Results of non-linear probing on VQA v2 dataset when using different scales of training data.}
\label{tbl:scale_vqa}
\end{table*}

\subsection{Image Encoder Architecture}

To evaluate the influence of encoder architectures on our methods, we pre-trained the models with different image encoder architectures. The results for various downstream tasks are reported in in Table \ref{tbl:enc_zs_cls}, \ref{tbl:enc_lp_cls}, \ref{tbl:enc_zs_itr}, and \ref{tbl:enc_vqa}. We also report the computation costs when using different encoder architectures in Table \ref{tbl:cost}.

\begin{table*}[t!]

\centering
\resizebox{0.7\linewidth}{!}{
\begin{tabular}{@{}ccccccccccc@{}}
\toprule
\multicolumn{1}{l}{} &
  \rotatebox{0}{C10} &
  \rotatebox{0}{C100} &
  \rotatebox{0}{F101} &
  \rotatebox{0}{PETS} &
  \rotatebox{0}{FLOW} &
  \rotatebox{0}{SUN} &
  \rotatebox{0}{DTD} &
  \rotatebox{0}{CAL} &
  \rotatebox{0}{IN} &
  \rotatebox{0}{AVG} \\ \midrule

ViT-B/32                     & 60.4 & 33.5 & 39.6 & 23.1 & 54.0 & 42.0 & 17.0 & 65.5 & 37.0                 & 41.3                  \\
ViT-B/32+FDT                & 67.7 & 39.9 & 42.9 & 25.8 & 55.5 & 45.5 & 26.5 & 69.6 & 39.3                 & 45.9 ($\uparrow$ 4.6) \\ \midrule
ViT-B/16                     & 64.6 & 32.1 & 49.7 & 25.7 & 59.7 & 43.4 & 21.3 & 67.9 &  42.1& 45.2                 \\
ViT-B/16+FDT                & 74.0 & 42.1 & 49.4 & 28.5 & 62.2 & 50.5 & 25.1 & 71.4 &                   45.6& 49.9 ($\uparrow$ 4.7)  \\ \midrule

SwinV2-B                     &  58.3& 23.3& 39.3& 20.0&  55.2& 40.1& 18.9& 62.1&     38.9& 39.6                  \\
SwinV2-B+FDT               & 58.9& 26.0& 44.7& 23.8& 55.4& 43.3& 21.4& 66.2& 42.3&   42.4 ($\uparrow$ 2.8)    \\ \bottomrule

\end{tabular}}
\vspace{-5pt}
\caption{Zero-shot image classification accuracy (\%) when using different image encoder architectures. The dataset names are abbreviated. C10/100 is CIFAR10/100. F101 is Food101. FLOW is Flowers. CAL is Caltech. IN is ImageNet-1K. ``AVG'' is the average accuracy over all datasets.}
\label{tbl:enc_zs_cls}

\end{table*}

\begin{table*}[t!]
\centering
\resizebox{0.7\linewidth}{!}{
\begin{tabular}{@{}cccccccccccc@{}}
\toprule
\multicolumn{1}{l}{} &
  \rotatebox{0}{C10} &
  \rotatebox{0}{C100} &
  \rotatebox{0}{F101} &
  \rotatebox{0}{PETS} &
  \rotatebox{0}{FLOW} &
  \rotatebox{0}{SUN} &
  \multicolumn{1}{l}{\rotatebox{0}{CARS}} &
  \rotatebox{0}{DTD} &
  \rotatebox{0}{CAL} &
  \rotatebox{0}{Air} &
  \rotatebox{0}{AVG} \\ \midrule
ViT-B/32                     & 88.3 & 68.6 & 72.1 & 72.5 & 92.6 & 69.5 & 29.8                 & 67.8 & 86.2 & 27.7                     & 67.5                  \\
ViT-B/32+FDT                & 89.1 & 71.2 & 74.4 & 73.0   & 93.4 & 70.8 & 31.4                 & 69.4 & 87.7 & 27.9                     & 68.8 ($\uparrow$ 1.3) \\ \midrule

ViT-B/16                & 89.2& 69.5& 80.3& 75.1& 95.9& 73.4& 33.4& 71.5& 88.3& 32.0& 68.8 \\ 

ViT-B/16+FDT                     &   89.3&  71.6& 82.3& 75.8& 96.1& 74.2&  34.0& 71.8& 88.6&  29.3& 71.3 ($\uparrow$ 2.5)                \\\midrule

SwinV2-B                     & 85.6&  65.1&  78.5&  71.4&  94.3& 72.3&                  30.8&  69.4&  85.9&  32.1& 68.5                  \\
SwinV2-B+FDT                &86.8 & 67.5 & 80.5 & 75.6 & 94.8& 73.1& 33.4&  72.7& 88.9&  34.0& 70.7   ($\uparrow$ 2.2)                   \\ \bottomrule
\end{tabular}}
\vspace{-5pt}
\caption{Linear probing image classification accuracy (\%) when using different image encoder architectures. The dataset names are abbreviated. C10/100 is CIFAR10/100. F101 is Food101. FLOW is Flowers. CAL is Caltech. Air is Aircraft. ``AVG'' is the average accuracy over all datasets.} 
\label{tbl:enc_lp_cls}

\end{table*}

\begin{table*}[t!]
\centering
\resizebox{0.95\linewidth}{!}{
\begin{tabular}{@{}lcccccccccccccc@{}}
\toprule
                & \multicolumn{7}{c}{Flickr30K}                                     & \multicolumn{7}{c}{MSCOCO}                                        \\ \midrule
 & \multicolumn{3}{c}{Image Retrieval} & \multicolumn{3}{c}{Text Retrieval} &  & \multicolumn{3}{c}{Image Retrieval} & \multicolumn{3}{c}{Text Retrieval} &  \\
                & R@1  & R@5  & R@10 & R@1  & R@5  & R@10 & rsum                    & R@1  & R@5  & R@10 & R@1  & R@5  & R@10 & rsum                    \\ \midrule
ViT-B/32        & 27.6 & 53.9 & 64.4 & 42.8 & 71.5 & 82.9 & 343.1                   & 15.9 & 36.7 & 47.8 & 24.8 & 49.8 & 61.8 & 236.8                   \\
ViT-B/32+FDT   & 32.6 & 58.6 & 68.5 & 51.0   & 78.3 & 87.5 & 376.5 ($\uparrow$ 33.4) & 19.4 & 40.8 & 51.9 & 29.6 & 55.3 & 66.1 & 263.1 ($\uparrow$ 26.3) \\ \midrule
ViT-B/16        & 35.3 & 60.6 & 71.7 & 50.5 & 81.1 & 88.6 & 387.8                   & 19.3 & 41.3 & 52.8 & 29.7 & 54.3 & 66.2 & 263.6                   \\
ViT-B/16+FDT & 41.6 & 67.5 & 76.9 & 60.8 & 86.1 & 92.6 & 425.5($\uparrow$ 37.7)  & 23.4 & 46.7 & 58.0   & 35.3 & 60.4 & 71.6 & 295.4($\uparrow$ 31.8)  \\ \midrule
SwinV2-B        & 30.5 & 56.8 & 67.8 & 48.5 & 77.7 & 86.8 & 368.1                   & 17.7 & 38.4 & 49.7 & 26.0   & 52.1 & 63.7 & 247.6                   \\
SwinV2-B+FDT & 39.6 & 65.2 & 74.9 & 57.9 & 85.7 & 92.2 & 415.5($\uparrow$ 47.4)  & 22.3 & 44.9 & 56.2 & 33.8 & 60.1 & 71.0   & 288.3($\uparrow$ 40.7)  \\ \bottomrule
\end{tabular}}
\vspace{-5pt}
\caption{Zero-shot image-text retrieval results on the Flickr30K and MSCOCO (5K) datasets when using different image encoder architectures.}
\label{tbl:enc_zs_itr}
\end{table*}

\begin{table*}[t!]
\centering
\resizebox{0.4\linewidth}{!}{
\begin{tabular}{@{}lcccc@{}}
\toprule
                & y/n  & number & other & overall               \\ \midrule
ViT-B/32        & 67.7 & 31.9   & 33.6  & 47.5                  \\
ViT-B/32 + FDT   & 67.8 & 34.6   & 39.6  & 50.6 ($\uparrow$ 3.1) \\ \midrule
ViT-B/16        & 69.0 & 33.2   & 36.0  & 49.2                  \\
ViT-B/16 + FDT & 72.0   & 37.6   & 42.9  & 54.3($\uparrow$ 5.1)  \\ \midrule
SwinV2-B        & 67.8 & 29.4   & 32.1  & 46.5                  \\
SwinV2-B + FDT & 68.6 & 34.5   & 41.0    & 51.6($\uparrow$ 5.1)  \\ \bottomrule
\end{tabular}
}
\vspace{-5pt}
\caption{Results of non-linear probing on VQA v2 dataset when using different image encoder architectures.}
\label{tbl:enc_vqa}
\end{table*}

\begin{table*}[t!]
\centering
\resizebox{0.5\linewidth}{!}{
\begin{tabular}{@{}lcccc@{}}
\toprule
 & \#param & FLOPs & \begin{tabular}[c]{@{}c@{}}Training time\\ (s/iter)\end{tabular} & \begin{tabular}[c]{@{}c@{}}Inference throughput\\ (image-text pairs/s)\end{tabular} \\ \midrule
CLIP-ViT-B/32     & 151M & 7.3G  & 0.50 &  808.5\\
CLIP-ViT-B/32+FDT & 161M & 9.4G  & 0.60 & 642.8  \\ \midrule
CLIP-ViT-B/16     & 150M & 20.5G & 1.15 & 315.7  \\
CLIP-ViT-B/16+FDT & 160M & 25.1G & 1.29 & 272.5  \\ \midrule
CLIP-Swin-B       & 151M & 18.4G & 1.41 & 258.3  \\
CLIP-Swin-B+FDT   & 161M & 20.5G & 1.51 & 248.1  \\ \bottomrule
\end{tabular}}
\vspace{-5pt}
\caption{Computation cost when using different image encoder architecture.}
\label{tbl:cost}
\end{table*}

\subsection{FDT Number}
The results of models trained with
different FDT numbers are shown in Table \ref{tbl:fdt_zs_cls}, \ref{tbl:fdt_lp_cls}, \ref{tbl:fdt_zs_itr}, and \ref{tbl:fdt_vqa}.

\begin{table*}[t!]
\centering
\resizebox{0.7\linewidth}{!}{
\begin{tabular}{@{}lcccccccccc@{}}
\toprule
FDT size &
  \rotatebox{0}{C10} &
  \rotatebox{0}{C100} &
  \rotatebox{0}{F101} &
  \rotatebox{0}{PETS} &
  \rotatebox{0}{FLOW} &
  \rotatebox{0}{SUN} &
  \rotatebox{0}{DTD} &
  \rotatebox{0}{CAL} &
  \rotatebox{0}{IN} &
  \rotatebox{0}{AVG} \\ \midrule
-     & 60.4          & 33.5          & 39.6          & 23.1          & 54.0 & 42.0          & 17.0          & 65.5          & 37.0          & 41.3          \\
8192  & \textbf{70.4} & \textbf{40.4} & 38.3          & 19.9          & 51.3 & 42.8          & 16.6          & 68.1          & 37.8          & 42.8          \\
16384 & 67.7          & 39.9          & \textbf{42.9} & \textbf{25.8} & 55.5 & \textbf{45.5} & \textbf{26.5} & 69.6          & 39.3          & \textbf{45.9} \\
24576 & 69.0          & 39.1          & 41.9          & 24.2          & \textbf{55.7} & 44.4          & 21.8          & \textbf{70.5} & \textbf{39.8} & 45.2          \\ \bottomrule
\end{tabular}}
\vspace{-5pt}
\caption{Zero-shot image classification accuracy (\%) of models  with different FDT sizes. The row whose FDT value is ``-'' represents the CLIP model. The dataset names are abbreviated. C10/100 is CIFAR10/100. F101 is Food101. FLOW is Flowers. CAL is Caltech. IN is ImageNet-1K. ``AVG'' is the average accuracy over all datasets.}
\label{tbl:fdt_zs_cls}
\end{table*}

\begin{table*}[t!]
\centering
\resizebox{0.7\linewidth}{!}{
\begin{tabular}{@{}cccccccccccc@{}}
\toprule
\multicolumn{1}{l}{FDT size} &
  \rotatebox{0}{C10} &
  \rotatebox{0}{C100} &
  \rotatebox{0}{F101} &
  \rotatebox{0}{PETS} &
  \rotatebox{0}{FLOW} &
  \rotatebox{0}{SUN} &
  \multicolumn{1}{l}{\rotatebox{0}{CARS}} &
  \rotatebox{0}{DTD} &
  \rotatebox{0}{CAL} &
  \rotatebox{0}{Air} &
  \rotatebox{0}{AVG} \\ \midrule
-                     & 88.3 & 68.6 & 72.1 & 72.5 & 92.6 & 69.5 & 29.8                 & 67.8 & 86.2 & 27.7                     & 67.5                  \\
8192  & 89.1 & 70.3 & 72.8 & 70.7 & \textbf{93.4} & 70.1 & 29.6 & 68.5 & 87.2 & 27.5 & 67.9 \\
16384 & 89.1 & \textbf{71.2} & 74.4 & \textbf{73.0} & \textbf{93.4} & \textbf{70.8} & \textbf{31.4} & 69.4 & \textbf{87.7} & 27.9 & \textbf{68.8} \\
24576 & \textbf{89.3} & 71.0 & \textbf{74.9} & 71.2 & \textbf{93.4} & 70.6 & 30.1 & \textbf{69.8} & 87.2 & \textbf{28.7} & 68.6                 \\ \bottomrule
\end{tabular}}
\vspace{-5pt}
\caption{Linear probing image classification accuracy (\%) of models  with different FDT sizes. The row whose FDT value is ``-'' represents the CLIP model. The dataset names are abbreviated. C10/100 is CIFAR10/100. F101 is Food101. FLOW is Flowers. CAL is Caltech. Air is Aircraft. ``AVG'' is the average accuracy over all datasets.}
\label{tbl:fdt_lp_cls}
\end{table*}

\begin{table*}[t!]
\centering
\resizebox{0.8\linewidth}{!}{
\begin{tabular}{@{}lcccccccccccccc@{}}
\toprule
         & \multicolumn{7}{c}{Flickr30K}                   & \multicolumn{7}{c}{MSCOCO}                      \\ \midrule
 &
  \multicolumn{3}{c}{Image Retrieval} &
  \multicolumn{3}{c}{Text Retrieval} &
   &
  \multicolumn{3}{c}{Image Retrieval} &
  \multicolumn{3}{c}{Text Retrieval} &
   \\
FDT size & R@1  & R@5  & R@10 & R@1  & R@5  & R@10 & rsum  & R@1  & R@5  & R@10 & R@1  & R@5  & R@10 & rsum  \\ \midrule
-        & 27.6 & 53.9 & 64.4 & 42.8 & 71.5 & 82.9 & 343.1 & 15.9 & 36.7 & 47.8 & 24.8 & 49.8 & 61.8 & 236.8 \\
8192     & 32.7 & 58.3 & 68.7 & 50.6 & 77.4 & 86.9 & 374.6 & 18.5 & 40.4 & 51.7 & 29.1 & 53.6 & 64.8 & 258.1 \\
16384 &
  32.6 &
  58.6 &
  68.5 &
  \textbf{51.0} &
  \textbf{78.3} &
  \textbf{87.5} &
  376.5 &
  \textbf{19.4} &
  \textbf{40.8} &
  \textbf{51.9} &
  29.6 &
  55.3 &
  66.1 &
  \textbf{263.1} \\
24576 &
  \textbf{33.3} &
  \textbf{60.3} &
  \textbf{70.4} &
  50.4 &
  78.1 &
  86.0 &
  \textbf{378.5} &
  18.6 &
  40.3 &
  51.8 &
  \textbf{29.7} &
  \textbf{55.8} &
  \textbf{66.9} &
  \textbf{263.1} \\ \bottomrule
\end{tabular}

}
\vspace{-5pt}
\caption{Zero-shot image-text retrieval results on the Flickr30K and MSCOCO (5K) datasets of models with different FDT sizes. The row whose FDT value is ``-'' represents the CLIP model.}
\label{tbl:fdt_zs_itr}
\end{table*}

\begin{table*}[t!]
\centering
\resizebox{0.3\linewidth}{!}{
\begin{tabular}{@{}lcccc@{}}
\toprule
FDT size & y/n           & number        & other         & overall       \\ \midrule
-        & 67.7          & 31.9          & 33.6          & 47.5          \\
8192     & 68.1          & 33.3          & 38.5          & 50.1          \\
16384    & 67.8          & 34.6          & 39.6          & 50.6          \\
24576    & \textbf{68.7} & \textbf{35.2} & \textbf{40.3} & \textbf{51.4} \\ \bottomrule
\end{tabular}}
\vspace{-5pt}
\caption{Results of non-linear probing on VQA v2 dataset of models with different FDT sizes. The row whose FDT value is ``-'' represents the CLIP model.}
\label{tbl:fdt_vqa}
\end{table*}

\subsection{Sparse Constraints}
We report the results of the models  trained with and without sparse constraint in Table \ref{tbl:sparse_zs_cls}, \ref{tbl:sparse_lp_cls}, \ref{tbl:sparse_zs_itr}, and \ref{tbl:sparse_vqa}.

\begin{table*}[t!]
\centering
\resizebox{0.8\linewidth}{!}{
\begin{tabular}{@{}lcccccccccc@{}}
\toprule
 &
  \rotatebox{0}{C10} &
  \rotatebox{0}{C100} &
  \rotatebox{0}{F101} &
  \rotatebox{0}{PETS} &
  \rotatebox{0}{FLOW} &
  \rotatebox{0}{SUN} &
  \rotatebox{0}{DTD} &
  \rotatebox{0}{CAL} &
  \rotatebox{0}{IN} &
  \rotatebox{0}{AVG} \\ \midrule
CLIP                              & 60.4 & 33.5 & 39.6 & 23.1 & 54.0 & 42.0 & 17.0 & 65.5 & 37.0 & 41.3 \\
CLIP+FDT$_{\mathrm{Softmax}}$ *   & 23.7 & 1.2  & 4.6  & 2.7  & 1.8  & 3.5  & 4.2  & 4.1  & 1.2  & 5.2  \\
CLIP+FDT$_{\mathrm{Sparsemax}}$ * & 59.9 & 24.7 & 17.3 & 20.9 & 35.1 & 31.2 & 20.8 & 56.8 & 25.0   & 32.4 \\ \midrule
CLIP+FDT$_{\mathrm{Softmax}}$     & 68.7 & 36.9 & 35.5 & 27.9 & 53.8 & 43.8 & 23.1 & 66.6 & 38.6 & 43.9 \\
CLIP+FDT$_{\mathrm{Sparsemax}}$   & 67.7 & 39.9 & 42.9 & 25.8 & 55.5 & 45.5 & 26.5 & 69.6 & 39.3 & 45.6 \\ \bottomrule
\end{tabular}}
\vspace{-5pt}
\caption{Zero-shot image classification accuracy (\%) of models trained with (Sparsemax) and without (Softmax) sparse constraints. The rows marked with ``*'' are the results when using FDT weights as features. The dataset names are abbreviated. C10/100 is CIFAR10/100. F101 is Food101. FLOW is Flowers. CAL is Caltech. IN is ImageNet-1K. ``AVG'' is the average accuracy over all datasets.}
\label{tbl:sparse_zs_cls}
\end{table*}

\begin{table*}[t!]
\centering
\resizebox{0.8\linewidth}{!}{
\begin{tabular}{@{}cccccccccccc@{}}
\toprule
\multicolumn{1}{l}{} &
  \rotatebox{0}{C10} &
  \rotatebox{0}{C100} &
  \rotatebox{0}{F101} &
  \rotatebox{0}{PETS} &
  \rotatebox{0}{FLOW} &
  \rotatebox{0}{SUN} &
  \multicolumn{1}{l}{\rotatebox{0}{CARS}} &
  \rotatebox{0}{DTD} &
  \rotatebox{0}{CAL} &
  \rotatebox{0}{Air} &
  \rotatebox{0}{AVG} \\ \midrule
CLIP                     & 88.3 & 68.6 & 72.1 & 72.5 & 92.6 & 69.5 & 29.8                 & 67.8 & 86.2 & 27.7                     & 67.5                  \\
CLIP+FDT$_{\mathrm{Softmax}}$  & 88.0 & 71.7 & 74.8 & 71.9 & 93.8 & 70.4 & 30.5 & 69.8 & 87.3 & 28.6 & 68.7 \\
CLIP+FDT$_{\mathrm{Sparsemax}}$ & 89.1 & 71.2 & 74.4 & 73.0   & 93.4 & 70.8 & 31.4 & 69.4 & 87.7 & 27.9 & 68.8 \\\bottomrule
\end{tabular}}
\vspace{-5pt}
\caption{Linear probing image classification accuracy (\%) of models  trained with (Sparsemax) and without (Softmax) sparse constraints. The dataset names are abbreviated. C10/100 is CIFAR10/100. F101 is Food101. FLOW is Flowers. CAL is Caltech. Air is Aircraft. ``AVG'' is the average accuracy over all datasets.}
\label{tbl:sparse_lp_cls}
\end{table*}

\begin{table*}[t!]
\centering
\resizebox{0.95\linewidth}{!}{
\begin{tabular}{@{}lcccccccccccccc@{}}
\toprule
                                  & \multicolumn{7}{c}{Flickr30K}                   & \multicolumn{7}{c}{MSCOCO}                      \\ \midrule
 & \multicolumn{3}{c}{Image Retrieval} & \multicolumn{3}{c}{Text Retrieval} &  & \multicolumn{3}{c}{Image Retrieval} & \multicolumn{3}{c}{Text Retrieval} &  \\
FDT size                          & R@1  & R@5  & R@10 & R@1  & R@5  & R@10 & rsum  & R@1  & R@5  & R@10 & R@1  & R@5  & R@10 & rsum  \\ \midrule
CLIP                              & 27.6 & 53.9 & 64.4 & 42.8 & 71.5 & 82.9 & 343.1 & 15.9 & 36.7 & 47.8 & 24.8 & 49.8 & 61.8 & 236.8 \\
CLIP+FDT$_{\mathrm{Softmax}}$ *   & 5.4  & 12.0 & 16.3 & 1.7  & 3.8  & 6.3  & 45.5  & 2.4  & 6.8  & 9.7  & 0.8  & 2.4  & 4.1  & 26.2  \\
CLIP+FDT$_{\mathrm{Sparsemax}}$ * & 10.5 & 29.8 & 39.2 & 32.5 & 59.8 & 70.6 & 242.4 & 6.0  & 16.5 & 24.1 & 18.3 & 40.5 & 52.1 & 157.5 \\ \midrule
CLIP+FDT$_{\mathrm{Softmax}}$     & 33.3 & 60.7 & 69.5 & 47.9 & 78.0   & 88.2 & 377.6 & 19.2 & 40.3 & 51.7 & 28.3 & 53.8 & 65.5 & 258.8 \\
CLIP+FDT$_{\mathrm{Sparsemax}}$   & 32.6 & 58.6 & 68.5 & 51.0 & 78.3 & 87.5 & 376.5 & 19.4 & 40.8 & 51.9 & 29.6 & 55.3 & 66.1 & 263.1 \\ \bottomrule
\end{tabular}}
\vspace{-5pt}
\caption{Zero-shot image-text retrieval results on the Flickr30K and MSCOCO (5K) datasets of models trained with (Sparsemax) and without (Softmax) sparse constraints. The rows marked with ``*'' are the results when using FDT weights as features.}
\label{tbl:sparse_zs_itr}

\end{table*}

\begin{table*}[t]
\centering
\resizebox{0.45\linewidth}{!}{
\begin{tabular}{@{}lcccc@{}}
\toprule
 & y/n           & number        & other         & overall       \\ \midrule
CLIP        & 67.7          & 31.9          & 33.6          & 47.5          \\
CLIP+FDT$_{\mathrm{Softmax}}$     & 65.7          & 31.9          & 36.2          & 47.9          \\
CLIP+FDT$_{\mathrm{Sparsemax}}$    & 67.8          & 34.6          & 39.6          & 50.6 \\ \bottomrule
\end{tabular}}
\vspace{-5pt}
\caption{Results of non-linear probing on VQAv2 dataset of models trained with (Sparsemax) and without (Softmax) sparse constraints.}
\label{tbl:sparse_vqa}
\end{table*}


%% file: main_paper.bbl
\begin{thebibliography}{10}\itemsep=-1pt

\bibitem{amir2021deep}
Shir Amir, Yossi Gandelsman, Shai Bagon, and Tali Dekel.
\newblock Deep vit features as dense visual descriptors.
\newblock {\em arXiv preprint arXiv:2112.05814}, 2(3):4, 2021.

\bibitem{vqa}
Stanislaw Antol, Aishwarya Agrawal, Jiasen Lu, Margaret Mitchell, Dhruv Batra,
  C~Lawrence Zitnick, and Devi Parikh.
\newblock Vqa: Visual question answering.
\newblock In {\em Proceedings of the IEEE international conference on computer
  vision}, pages 2425--2433, 2015.

\bibitem{mm_dict_cls}
Soheil Bahrampour, Nasser~M Nasrabadi, Asok Ray, and William~Kenneth Jenkins.
\newblock Multimodal task-driven dictionary learning for image classification.
\newblock {\em IEEE transactions on Image Processing}, 25(1):24--38, 2015.

\bibitem{food101}
Lukas Bossard, Matthieu Guillaumin, and Luc~Van Gool.
\newblock Food-101--mining discriminative components with random forests.
\newblock In {\em European conference on computer vision}, pages 446--461.
  Springer, 2014.

\bibitem{cc12m}
Soravit Changpinyo, Piyush Sharma, Nan Ding, and Radu Soricut.
\newblock Conceptual 12m: Pushing web-scale image-text pre-training to
  recognize long-tail visual concepts.
\newblock In {\em Proceedings of the IEEE/CVF Conference on Computer Vision and
  Pattern Recognition}, pages 3558--3568, 2021.

\bibitem{chen2018you}
Tianlang Chen, Yuxiao Chen, Han Guo, and Jiebo Luo.
\newblock You type a few words and we do the rest: Image recommendation for
  social multimedia posts.
\newblock In {\em 2018 IEEE International Conference on Big Data (Big Data)},
  pages 2124--2133. IEEE, 2018.

\bibitem{chen2023more}
Yuxiao Chen, Jianbo Yuan, Long Zhao, Tianlang Chen, Rui Luo, Larry Davis, and
  Dimitris~N Metaxas.
\newblock More than just attention: Improving cross-modal attentions with
  contrastive constraints for image-text matching.
\newblock In {\em Proceedings of the IEEE/CVF Winter Conference on Applications
  of Computer Vision}, pages 4432--4440, 2023.

\bibitem{dtd}
Mircea Cimpoi, Subhransu Maji, Iasonas Kokkinos, Sammy Mohamed, and Andrea
  Vedaldi.
\newblock Describing textures in the wild.
\newblock In {\em Proceedings of the IEEE conference on computer vision and
  pattern recognition}, pages 3606--3613, 2014.

\bibitem{declip_benchmark}
Yufeng Cui, Lichen Zhao, Feng Liang, Yangguang Li, and Jing Shao.
\newblock Democratizing contrastive language-image pre-training: A clip
  benchmark of data, model, and supervision.
\newblock {\em arXiv preprint arXiv:2203.05796}, 2022.

\bibitem{imagenet}
Jia Deng, Wei Dong, Richard Socher, Li-Jia Li, Kai Li, and Li Fei-Fei.
\newblock Imagenet: A large-scale hierarchical image database.
\newblock In {\em 2009 IEEE conference on computer vision and pattern
  recognition}, pages 248--255. Ieee, 2009.

\bibitem{bert}
Jacob Devlin, Ming-Wei Chang, Kenton Lee, and Kristina Toutanova.
\newblock Bert: Pre-training of deep bidirectional transformers for language
  understanding.
\newblock {\em arXiv preprint arXiv:1810.04805}, 2018.

\bibitem{vit}
Alexey Dosovitskiy, Lucas Beyer, Alexander Kolesnikov, Dirk Weissenborn,
  Xiaohua Zhai, Thomas Unterthiner, Mostafa Dehghani, Matthias Minderer, Georg
  Heigold, Sylvain Gelly, Jakob Uszkoreit, and Neil Houlsby.
\newblock An image is worth 16x16 words: Transformers for image recognition at
  scale.
\newblock In {\em International Conference on Learning Representations}, 2021.

\bibitem{meter}
Zi-Yi Dou, Yichong Xu, Zhe Gan, Jianfeng Wang, Shuohang Wang, Lijuan Wang,
  Chenguang Zhu, Pengchuan Zhang, Lu Yuan, Nanyun Peng, et~al.
\newblock An empirical study of training end-to-end vision-and-language
  transformers.
\newblock In {\em Proceedings of the IEEE/CVF Conference on Computer Vision and
  Pattern Recognition}, pages 18166--18176, 2022.

\bibitem{caltech}
Li Fei-Fei, Rob Fergus, and Pietro Perona.
\newblock Learning generative visual models from few training examples: An
  incremental bayesian approach tested on 101 object categories.
\newblock In {\em 2004 conference on computer vision and pattern recognition
  workshop}, pages 178--178. IEEE, 2004.

\bibitem{mm_dict_conv}
Fangyuan Gao, Xin Deng, Mai Xu, Jingyi Xu, and Pier~Luigi Dragotti.
\newblock Multi-modal convolutional dictionary learning.
\newblock {\em IEEE Transactions on Image Processing}, 31:1325--1339, 2022.

\bibitem{gao2021clip}
Peng Gao, Shijie Geng, Renrui Zhang, Teli Ma, Rongyao Fang, Yongfeng Zhang,
  Hongsheng Li, and Yu Qiao.
\newblock Clip-adapter: Better vision-language models with feature adapters.
\newblock {\em arXiv preprint arXiv:2110.04544}, 2021.

\bibitem{conv_dict}
Cristina Garcia-Cardona and Brendt Wohlberg.
\newblock Convolutional dictionary learning: A comparative review and new
  algorithms.
\newblock {\em IEEE Transactions on Computational Imaging}, 4(3):366--381,
  2018.

\bibitem{geng2023hiclip}
Shijie Geng, Jianbo Yuan, Yu Tian, Yuxiao Chen, and Yongfeng Zhang.
\newblock Hi{CLIP}: Contrastive language-image pretraining with hierarchy-aware
  attention.
\newblock In {\em The Eleventh International Conference on Learning
  Representations}, 2023.

\bibitem{resnet}
Kaiming He, Xiangyu Zhang, Shaoqing Ren, and Jian Sun.
\newblock Deep residual learning for image recognition.
\newblock In {\em Proceedings of the IEEE conference on computer vision and
  pattern recognition}, pages 770--778, 2016.

\bibitem{learning_concepts}
Yan Huang, Qi Wu, Chunfeng Song, and Liang Wang.
\newblock Learning semantic concepts and order for image and sentence matching.
\newblock In {\em Proceedings of the IEEE Conference on Computer Vision and
  Pattern Recognition}, pages 6163--6171, 2018.

\bibitem{soho}
Zhicheng Huang, Zhaoyang Zeng, Yupan Huang, Bei Liu, Dongmei Fu, and Jianlong
  Fu.
\newblock Seeing out of the box: End-to-end pre-training for vision-language
  representation learning.
\newblock In {\em Proceedings of the IEEE/CVF Conference on Computer Vision and
  Pattern Recognition}, pages 12976--12985, 2021.

\bibitem{align}
Chao Jia, Yinfei Yang, Ye Xia, Yi-Ting Chen, Zarana Parekh, Hieu Pham, Quoc Le,
  Yun-Hsuan Sung, Zhen Li, and Tom Duerig.
\newblock Scaling up visual and vision-language representation learning with
  noisy text supervision.
\newblock In {\em International Conference on Machine Learning}, pages
  4904--4916. PMLR, 2021.

\bibitem{cars}
Jonathan Krause, Michael Stark, Jia Deng, and Li Fei-Fei.
\newblock 3d object representations for fine-grained categorization.
\newblock In {\em Proceedings of the IEEE international conference on computer
  vision workshops}, pages 554--561, 2013.

\bibitem{cifar10}
Alex Krizhevsky, Geoffrey Hinton, et~al.
\newblock Learning multiple layers of features from tiny images.
\newblock 2009.

\bibitem{li2022blip}
Junnan Li, Dongxu Li, Caiming Xiong, and Steven Hoi.
\newblock Blip: Bootstrapping language-image pre-training for unified
  vision-language understanding and generation.
\newblock {\em arXiv preprint arXiv:2201.12086}, 2022.

\bibitem{ALBEF}
Junnan Li, Ramprasaath~R. Selvaraju, Akhilesh~Deepak Gotmare, Shafiq Joty,
  Caiming Xiong, and Steven Hoi.
\newblock Align before fuse: Vision and language representation learning with
  momentum distillation.
\newblock In {\em NeurIPS}, 2021.

\bibitem{li2022unimo}
Wei Li, Can Gao, Guocheng Niu, Xinyan Xiao, Hao Liu, Jiachen Liu, Hua Wu, and
  Haifeng Wang.
\newblock Unimo-2: End-to-end unified vision-language grounded learning.
\newblock {\em arXiv preprint arXiv:2203.09067}, 2022.

\bibitem{declip}
Yangguang Li, Feng Liang, Lichen Zhao, Yufeng Cui, Wanli Ouyang, Jing Shao,
  Fengwei Yu, and Junjie Yan.
\newblock Supervision exists everywhere: A data efficient contrastive
  language-image pre-training paradigm.
\newblock {\em arXiv preprint arXiv:2110.05208}, 2021.

\bibitem{mscoco}
Tsung-Yi Lin, Michael Maire, Serge Belongie, James Hays, Pietro Perona, Deva
  Ramanan, Piotr Doll{\'a}r, and C~Lawrence Zitnick.
\newblock Microsoft coco: Common objects in context.
\newblock In {\em European conference on computer vision}, pages 740--755.
  Springer, 2014.

\bibitem{lin2022frozen}
Ziyi Lin, Shijie Geng, Renrui Zhang, Peng Gao, Gerard de Melo, Xiaogang Wang,
  Jifeng Dai, Yu Qiao, and Hongsheng Li.
\newblock Frozen clip models are efficient video learners.
\newblock In {\em Computer Vision--ECCV 2022: 17th European Conference, Tel
  Aviv, Israel, October 23--27, 2022, Proceedings, Part XXXV}, pages 388--404.
  Springer, 2022.

\bibitem{crossdiscrete}
Alexander~H Liu, SouYoung Jin, Cheng-I~Jeff Lai, Andrew Rouditchenko, Aude
  Oliva, and James Glass.
\newblock Cross-modal discrete representation learning.
\newblock {\em arXiv preprint arXiv:2106.05438}, 2021.

\bibitem{cosinedecay}
Ilya Loshchilov and Frank Hutter.
\newblock Sgdr: Stochastic gradient descent with warm restarts.
\newblock {\em arXiv preprint arXiv:1608.03983}, 2016.

\bibitem{loshchilov2018decoupled}
Ilya Loshchilov and Frank Hutter.
\newblock Decoupled weight decay regularization.
\newblock In {\em International Conference on Learning Representations}, 2019.

\bibitem{aircraft}
Subhransu Maji, Esa Rahtu, Juho Kannala, Matthew Blaschko, and Andrea Vedaldi.
\newblock Fine-grained visual classification of aircraft.
\newblock {\em arXiv preprint arXiv:1306.5151}, 2013.

\bibitem{sparsemax}
Andre Martins and Ramon Astudillo.
\newblock From softmax to sparsemax: A sparse model of attention and
  multi-label classification.
\newblock In {\em International conference on machine learning}, pages
  1614--1623. PMLR, 2016.

\bibitem{slip}
Norman Mu, Alexander Kirillov, David Wagner, and Saining Xie.
\newblock Slip: Self-supervision meets language-image pre-training.
\newblock {\em arXiv preprint arXiv:2112.12750}, 2021.

\bibitem{flower}
Maria-Elena Nilsback and Andrew Zisserman.
\newblock Automated flower classification over a large number of classes.
\newblock In {\em 2008 Sixth Indian Conference on Computer Vision, Graphics \&
  Image Processing}, pages 722--729. IEEE, 2008.

\bibitem{pets}
Omkar~M Parkhi, Andrea Vedaldi, Andrew Zisserman, and CV Jawahar.
\newblock Cats and dogs.
\newblock In {\em 2012 IEEE conference on computer vision and pattern
  recognition}, pages 3498--3505. IEEE, 2012.

\bibitem{peng2022beit2}
Zhiliang Peng, Li Dong, Hangbo Bao, Qixiang Ye, and Furu Wei.
\newblock Beit v2: Masked image modeling with vector-quantized visual
  tokenizers.
\newblock {\em arXiv preprint arXiv:2208.06366}, 2022.

\bibitem{clip}
Alec Radford, Jong~Wook Kim, Chris Hallacy, Aditya Ramesh, Gabriel Goh,
  Sandhini Agarwal, Girish Sastry, Amanda Askell, Pamela Mishkin, Jack Clark,
  et~al.
\newblock Learning transferable visual models from natural language
  supervision.
\newblock In {\em International Conference on Machine Learning}, pages
  8748--8763. PMLR, 2021.

\bibitem{gpt}
Alec Radford, Karthik Narasimhan, Tim Salimans, Ilya Sutskever, et~al.
\newblock Improving language understanding by generative pre-training.
\newblock 2018.

\bibitem{Gra_unif}
Zhiyin Shao, Xinyu Zhang, Meng Fang, Zhifeng Lin, Jian Wang, and Changxing
  Ding.
\newblock Learning granularity-unified representations for text-to-image person
  re-identification.
\newblock In {\em Proceedings of the 30th ACM International Conference on
  Multimedia}, pages 5566--5574, 2022.

\bibitem{cc3m}
Piyush Sharma, Nan Ding, Sebastian Goodman, and Radu Soricut.
\newblock Conceptual captions: A cleaned, hypernymed, image alt-text dataset
  for automatic image captioning.
\newblock In {\em Proceedings of the 56th Annual Meeting of the Association for
  Computational Linguistics (Volume 1: Long Papers)}, pages 2556--2565, 2018.

\bibitem{singh2022flava}
Amanpreet Singh, Ronghang Hu, Vedanuj Goswami, Guillaume Couairon, Wojciech
  Galuba, Marcus Rohrbach, and Douwe Kiela.
\newblock {FLAVA:} {A} foundational language and vision alignment model.
\newblock In {\em CVPR}, 2022.

\bibitem{vqvae}
Aaron Van Den~Oord, Oriol Vinyals, et~al.
\newblock Neural discrete representation learning.
\newblock {\em Advances in neural information processing systems}, 30, 2017.

\bibitem{wang2022beit3}
Wenhui Wang, Hangbo Bao, Li Dong, Johan Bjorck, Zhiliang Peng, Qiang Liu, Kriti
  Aggarwal, Owais~Khan Mohammed, Saksham Singhal, Subhojit Som, et~al.
\newblock Image as a foreign language: Beit pretraining for all vision and
  vision-language tasks.
\newblock {\em arXiv preprint arXiv:2208.10442}, 2022.

\bibitem{sun397}
Jianxiong Xiao, James Hays, Krista~A Ehinger, Aude Oliva, and Antonio Torralba.
\newblock Sun database: Large-scale scene recognition from abbey to zoo.
\newblock In {\em 2010 IEEE computer society conference on computer vision and
  pattern recognition}, pages 3485--3492. IEEE, 2010.

\bibitem{image_infinity}
Lingxi Xie, Xiaopeng Zhang, Longhui Wei, Jianlong Chang, and Qi Tian.
\newblock What is considered complete for visual recognition?
\newblock {\em arXiv preprint arXiv:2105.13978}, 2021.

\bibitem{tcl}
Jinyu Yang, Jiali Duan, Son Tran, Yi Xu, Sampath Chanda, Liqun Chen, Belinda
  Zeng, Trishul Chilimbi, and Junzhou Huang.
\newblock Vision-language pre-training with triple contrastive learning.
\newblock In {\em Proceedings of the IEEE/CVF Conference on Computer Vision and
  Pattern Recognition}, pages 15671--15680, 2022.

\bibitem{yang2022vision}
Jinyu Yang, Jiali Duan, Son Tran, Yi Xu, Sampath Chanda, Liqun Chen, Belinda
  Zeng, Trishul Chilimbi, and Junzhou Huang.
\newblock Vision-language pre-training with triple contrastive learning.
\newblock 2022.

\bibitem{filip}
Lewei Yao, Runhui Huang, Lu Hou, Guansong Lu, Minzhe Niu, Hang Xu, Xiaodan
  Liang, Zhenguo Li, Xin Jiang, and Chunjing Xu.
\newblock Filip: Fine-grained interactive language-image pre-training.
\newblock {\em arXiv preprint arXiv:2111.07783}, 2021.

\bibitem{msclip}
Haoxuan You, Luowei Zhou, Bin Xiao, Noel Codella, Yu Cheng, Ruochen Xu, Shih-Fu
  Chang, and Lu Yuan.
\newblock Learning visual representation from modality-shared contrastive
  language-image pre-training.
\newblock In {\em European Conference on Computer Vision}, pages 69--87.
  Springer, 2022.

\bibitem{f30k}
Peter Young, Alice Lai, Micah Hodosh, and Julia Hockenmaier.
\newblock From image descriptions to visual denotations: New similarity metrics
  for semantic inference over event descriptions.
\newblock {\em Transactions of the Association for Computational Linguistics},
  2:67--78, 2014.

\bibitem{zhang2023prompt}
Renrui Zhang, Xiangfei Hu, Bohao Li, Siyuan Huang, Hanqiu Deng, Hongsheng Li,
  Yu Qiao, and Peng Gao.
\newblock Prompt, generate, then cache: Cascade of foundation models makes
  strong few-shot learners.
\newblock In {\em Proceedings of the IEEE/CVF Conference on Computer Vision and
  Pattern Recognition}, 2023.

\bibitem{zhang2022learning}
Renrui Zhang, Liuhui Wang, Yu Qiao, Peng Gao, and Hongsheng Li.
\newblock Learning 3d representations from 2d pre-trained models via
  image-to-point masked autoencoders.
\newblock In {\em Proceedings of the IEEE/CVF Conference on Computer Vision and
  Pattern Recognition}, 2023.

\bibitem{zhang2023parameter}
Renrui Zhang, Liuhui Wang, Yali Wang, Peng Gao, Hongsheng Li, and Jianbo Shi.
\newblock Parameter is not all you need: Starting from non-parametric networks
  for 3d point cloud analysis.
\newblock In {\em Proceedings of the IEEE/CVF Conference on Computer Vision and
  Pattern Recognition}, 2023.

\bibitem{zhang2022sine}
Zhixing Zhang, Ligong Han, Arnab Ghosh, Dimitris Metaxas, and Jian Ren.
\newblock Sine: Single image editing with text-to-image diffusion models.
\newblock {\em arXiv preprint arXiv:2212.04489}, 2022.

\end{thebibliography}
